\documentclass[11pt]{article}
\usepackage{iftex}
\ifPDFTeX
\fi

\usepackage[final]{acl}
\usepackage{times}
\usepackage{latexsym}
\usepackage[T1]{fontenc}
\usepackage[utf8]{inputenc}
\usepackage{inconsolata}

\usepackage{microtype}
\usepackage{graphicx}
\usepackage{url}
\usepackage{booktabs}
\usepackage{amsmath}
\usepackage{wrapfig}
\usepackage{algorithm}
\usepackage{algorithmicx}
\usepackage{algpseudocode}
\usepackage{multirow}

\usepackage[table]{xcolor}
\usepackage[most]{tcolorbox}
\usepackage{enumitem}
\definecolor{myblue}{RGB}{35,85,255}
\definecolor{mygray}{gray}{0.45}
\newcommand{\algblue}[1]{{\color{myblue}#1}}
\newcommand{\algphase}[1]{\textcolor{mygray}{\#~#1}}

\usepackage{amssymb}
\usepackage{subcaption}

\title{STRIDE: A Self-Reflective Agent Framework for Reliable Automatic Equation Discovery}

\author{
Jiarui Su$^{1}$ \quad Songjun Tu$^{3,2}$* \quad Bei Sun$^{1}$* \quad Xiaojun Liang$^{2}$ \\
$^{1}$Central South University \quad
$^{2}$Pengcheng Laboratory \\
$^{3}$Institute of Automation, Chinese Academy of Sciences
}

\begin{document}
\maketitle
\thispagestyle{plain}
\begingroup
\renewcommand{\thefootnote}{\fnsymbol{footnote}}

\footnotetext[1]{\noindent Corresponding author. \\ Emails: sujiarui2024@csu.edu.cn, tusongjun2023@ia.ac.cn, sunbei@csu.edu.cn, liangxj@pcl.ac.cn.}

\endgroup

\begin{abstract}
LLM-based equation discovery offers a promising route to recovering symbolic laws from data, but many systems still rely on generation-centered loops that propose candidates, fit parameters, score results, and reuse selected examples. Such loops can misjudge useful skeletons under unreliable fitting, discard near-correct equations that require repair, and accumulate redundant memories that provide limited guidance. We propose \textbf{STRIDE}, a self-reflective agent framework that improves reliability by coordinating data-aware generation, mixed-fitting evaluation, critic--executor repair, and diversity-preserving semantic memory. By turning fitted scores and candidate behavior into shared feedback, STRIDE enables equations to be proposed, assessed, refined, and reused within a closed-loop discovery process. Experiments on representative symbolic-regression benchmarks and LSR-Synth suites show that STRIDE improves accuracy, OOD robustness, and structural recovery across multiple LLM backbones, with ablations and analyses confirming the contribution of its core components. Code is available at: \url{https://github.com/jiaruisu0218-cloud/STRIDE}.
\end{abstract}

\begin{figure*}[t]
\centering
\includegraphics[width=\textwidth]{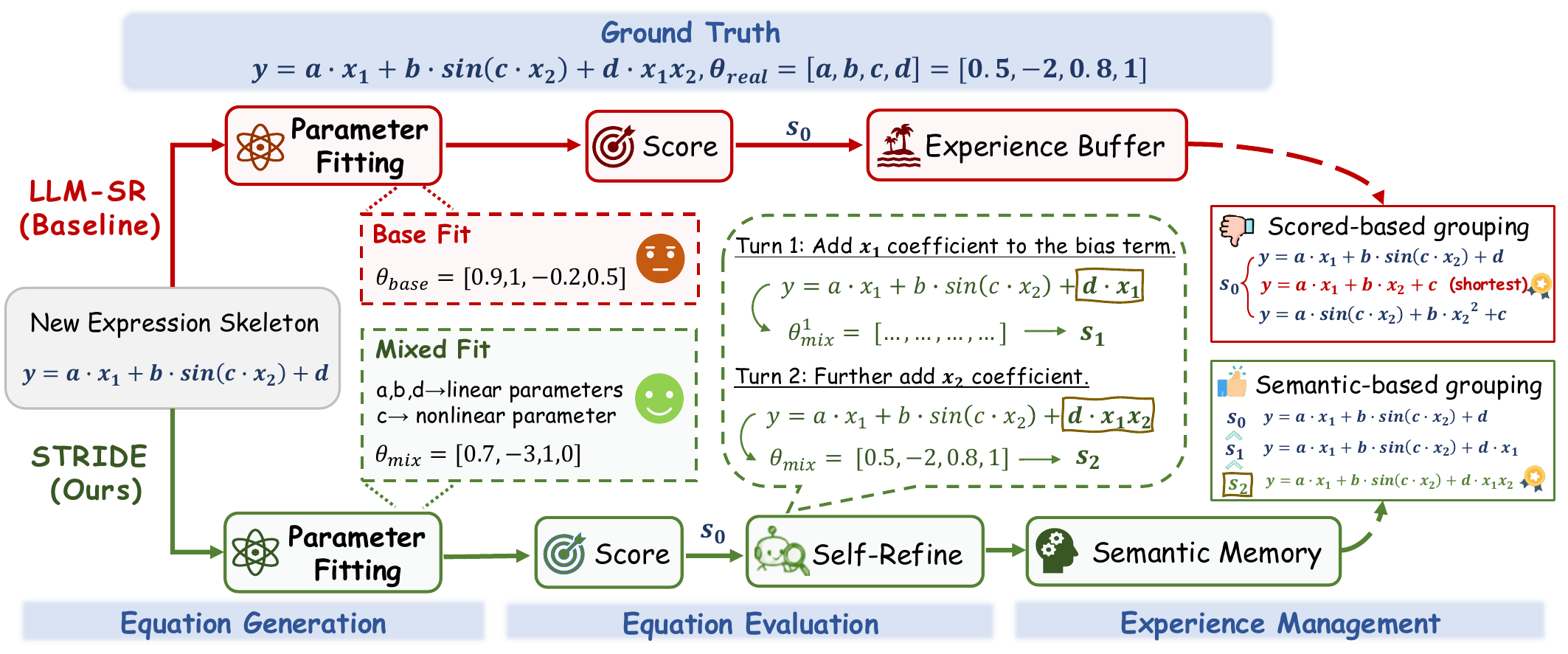}
\caption{STRIDE versus LLM-SR. The upper part shows a representative LLM-SR iteration that samples a skeleton, generates a candidate, evaluates it, and updates the library. STRIDE adds data-aware sampling, mixed fitting, reflective refinement, and semantic memory to repair promising candidates and discover more reliable equations.}
\label{fig:motivation}
\end{figure*}

\section{Introduction}
\label{sec:intro}
Equation discovery seeks compact, generalizable, and interpretable mathematical laws directly from data \citep{lemos2023rediscovering,makke2024review,dong2025recent}. Symbolic regression (SR) addresses this goal by searching over both equation structures and numerical parameters \citep{schmidt2009distilling}. Yet the symbolic search space grows combinatorially, and finite observations often admit spurious fits \citep{virgolin2022nphard,lacava2021contemporary,matsubara2024rethinking}. Reliable equation discovery therefore requires not only low training error, but also compact structure and robustness beyond the observed regime.

LLM-based equation discovery has recently emerged as a promising direction, leveraging LLM priors over mathematical syntax and scientific patterns \citep{shojaee2025llmsr,grayeli2024lasr,guo2025srllm,zhang-etal-2024-comprehensive-survey}. A common paradigm organizes discovery as repeated library-conditioned iterations: sample an expression skeleton from an experience library, ask an LLM to instantiate or modify it into a candidate equation, fit and score the candidate, and write selected cases back to the library for later prompting. This process expands the search beyond hand-crafted templates, but its reliability depends on more than generating plausible equations.

As illustrated in Figure~\ref{fig:motivation}, this generation-centered paradigm exposes a deeper coordination problem. A single generation step and a score-based library are expected to propose structures, assess their potential after fitting, decide whether imperfect candidates deserve repair, and preserve useful experience for later iterations. As a result, useful skeletons may be discarded when fitting underestimates their potential; near-correct equations may be abandoned rather than locally repaired; and complex but informative hypotheses may be lost under short-term fitness or length bias. These limitations suggest that reliable LLM-based equation discovery should be organized as a multi-role reflective agent workflow, where generation, evaluation, critique, repair, and memory updating are distinct but coordinated roles.

To address these limitations, we propose \textbf{STRIDE}, a \textbf{S}elf-reflective agen\textbf{T} Framework for \textbf{R}el\textbf{I}able automatic equation \textbf{D}iscov\textbf{E}ry. 
STRIDE instantiates this idea as a multi-role reflective workflow. \textbf{(I)} A generator agent conditions on task context, data hints, and retrieved memory cases to propose candidate skeletons. \textbf{(II)} A mixed-fitting evaluator fits free parameters and returns score feedback, giving the agents a more faithful view of skeleton quality. \textbf{(III)} A critic agent diagnoses promising but imperfect candidates from fitted-parameter feedback, while an executor agent turns the critic's actions into revised executable equations. \textbf{(IV)} A semantic memory module retains diverse high-scoring hypotheses and supplies them as experience for later agent iterations.

The main contributions are as follows:
\begin{tcolorbox}[
    enhanced,
    colback=black!2,
    colframe=black!15,
    boxrule=0.35pt,
    arc=1.5mm,
    sharp corners=southwest,
    sharp corners=southeast,
    left=1.2mm,right=1.2mm,top=1.2mm,bottom=1.2mm
]
\begin{enumerate}[leftmargin=1.5em, itemsep=2pt, topsep=2pt]
    \item We identify LLM-based equation discovery as a \textbf{multi-role self-reflective agent workflow} and introduce \textbf{STRIDE}, a coordinated framework for reliable equation discovery.
    \item We develop four reliability-oriented mechanisms, namely \textbf{data-aware sampling}, \textbf{mixed fitting}, \textbf{reflective refinement}, and \textbf{semantic memory}, to better propose, assess, repair, and retain candidate equations.
    \item We provide empirical evidence across benchmarks and LLM backbones, showing that STRIDE improves \textbf{reliability} against strong baselines in \textbf{ID/OOD} settings, with analyses validating each component.
\end{enumerate}
\end{tcolorbox}

\begin{figure*}[t]
\centering
\includegraphics[width=\textwidth]{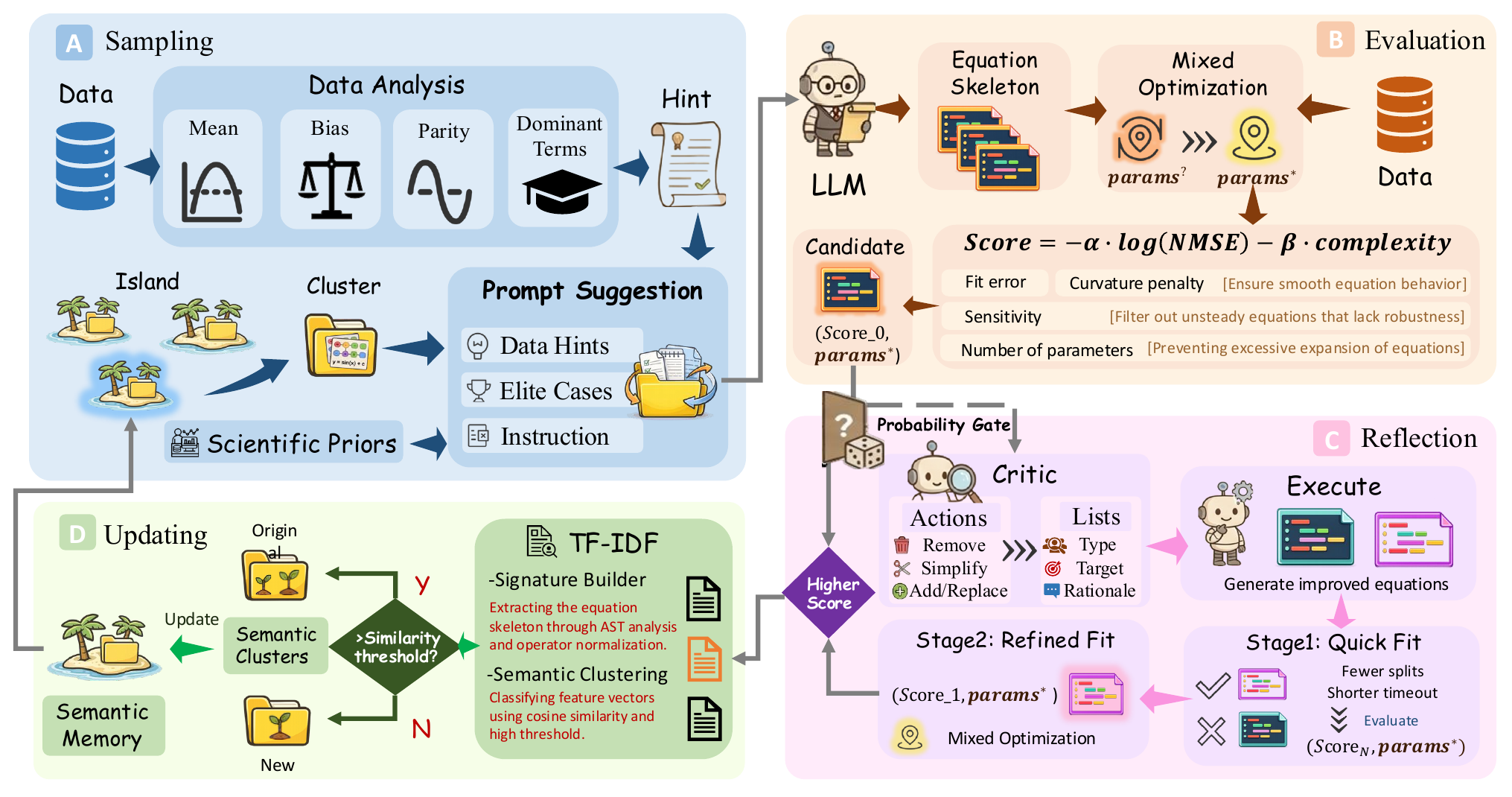}
\caption{Overall framework of STRIDE: data-aware generation, mixed-fitting evaluation, critic--executor repair, and semantic memory form a closed-loop equation discovery workflow.}
\label{fig:framework}
\end{figure*}

\section{Preliminaries}
\label{sec:problem}
\paragraph{Symbolic Regression.}
Given observed data
$\mathcal{D}=\{(\mathbf{x}_i,y_i)\}_{i=1}^{N},$
symbolic regression seeks an interpretable equation
$f(\mathbf{x};\boldsymbol{\theta})$
that predicts $y$ from $\mathbf{x}$. We formulate this task as a search problem over a symbolic hypothesis space
$\mathcal{F},$
where each candidate equation is composed from an operator set
$\mathcal{O}$
(e.g., $+,-,\times,\div,\sin,\cos,\exp$ and $\log$), input variables
$\mathbf{x},$
numerical constants, and free parameters
$\boldsymbol{\theta}.$
The goal is to identify a symbolic structure that explains the observations accurately while remaining compact, stable, and human-interpretable.

\paragraph{LLM-Based Equation Discovery.}
Recent LLM-based equation discovery methods, such as \textsc{LLM-SR}~\citep{shojaee2025llmsr}, use LLMs as scientific priors for symbolic regression. The upper part of Figure~\ref{fig:motivation} illustrates one representative iteration rather than the whole system. The step begins by sampling a new expression skeleton from an experience library. The skeleton is combined with the task description, variable information, and operator constraints to prompt an LLM, which instantiates or modifies it into an executable candidate equation or program body. An evaluator then fits the free parameters of this candidate, scores it by prediction error such as normalized mean squared error (NMSE) together with structural preferences, and selectively writes useful cases back to the library.

This paradigm turns equation discovery into a repeated process of library-conditioned sampling, LLM generation, parameter fitting, scoring, and library updating. It provides the functional basis for a role-based workflow: a generator proposes candidate equations, an evaluator fits and scores them, and a memory component retains selected experience for future iterations. Its goal is to find a reliable high-scoring equation
\begin{equation}
f^{*}
=
\arg\max_{f\in\mathcal{F}}
\mathrm{Score}(f,\boldsymbol{\theta}^{*};\mathcal{D}),
\label{eq:main_problem}
\end{equation}
where $\mathcal{F}$ is the candidate equation space defined by the available variables, operators, constants, and expression constraints; $f$ is a candidate symbolic equation; and $f^{*}$ is the selected best equation under the scoring objective. For each candidate $f$, $\boldsymbol{\theta}^{*}$ denotes the parameters fitted on the observed dataset $\mathcal{D}$, and $\mathrm{Score}(f,\boldsymbol{\theta}^{*};\mathcal{D})$ measures its quality by combining predictive fidelity with structural preference. The $\arg\max$ operator therefore selects the candidate with the highest score after parameter fitting. This formulation provides the basis for STRIDE, which augments the standard generation, evaluation, and memory loop with critic and executor agents for reflective repair.

\section{Method}
\label{sec:method}

\subsection{Overview}
Figures~\ref{fig:motivation} and~\ref{fig:framework} summarize STRIDE as a reflective workflow that couples data-aware generation, mixed parameter fitting, critic--executor repair, and semantic memory updating. In each iteration, STRIDE proposes parameterized skeletons, fits and scores them, optionally repairs promising candidates, and writes the selected structure back to memory for later sampling.

\subsection{Generator Agent: Data-Aware Sampling}
\label{sec:method_data}

Sampling is handled by the generator agent $\mathcal{M}_{\mathrm{gen}}$. 
STRIDE constructs the sampling prompt from three sources. First, it extracts a data hint $h$ from the training data $\mathcal{D}$ through lightweight data analysis, including mean statistics, bias tendency, parity patterns, and dominant terms from polynomial, interaction, and simple nonlinear feature expansions. Second, it retrieves elite cases $\mathcal{E}$ from the island and cluster structure of semantic memory $\mathcal{B}$. Third, it adds task instructions derived from the task specification $\mathcal{T}$ and variable descriptions.

Let $p$ denote the resulting prompt. Conditioned on $p$, the generator agent produces a batch of executable equation skeletons
\begin{equation}
\mathcal{C}=\{f_j\}_{j=1}^{b}, \qquad
f_j \sim \mathcal{M}_{\mathrm{gen}}(\cdot \mid p).
\end{equation}
Each skeleton contains free numerical parameters and is passed to the evaluator for fitting and scoring. This stage supplies data-aware structural hypotheses for the subsequent evaluation, repair, and memory-update steps; implementation cases are provided in Appendix~\ref{sec:effect_data}.
The generator is therefore responsible for structural proposal, while numerical validation is deferred to the evaluator.

\subsection{Evaluator: Mixed Parameter Fitting}
\label{sec:method_mixedopt}
The evaluator fits each generated skeleton and returns fitted parameters together with a score. Given a candidate equation \(f(\mathbf{u};\theta)\), STRIDE parses its skeleton into an abstract syntax tree (AST)~\citep{aho2006compilers} and separates the parameters into linear coefficients $\mathbf{w}$ and nonlinear or coupled parameters $\mathbf{q}$ according to their roles in the expression.

The evaluator then performs nested mixed fitting. The outer loop searches over nonlinear parameters $\mathbf{q}$, and the inner loop solves the best linear coefficients $\mathbf{w}^{\star}(\mathbf{q})$ for each proposed $\mathbf{q}$ and evaluates the resulting normalized error:
\begin{equation}
\begin{aligned}
\mathbf{w}^{\star}(\mathbf{q})
&= \arg\min_{\mathbf{w}}
\left\|f(\mathbf{u};\mathbf{w},\mathbf{q})-\mathbf{y}\right\|_2^2
+\lambda\|\mathbf{w}\|_2^2,\\
J(\mathbf{q})
&= \frac{1}{N\,\mathrm{Var}(\mathbf{y})}
\left\|f(\mathbf{u};\mathbf{w}^{\star}(\mathbf{q}),\mathbf{q})-\mathbf{y}\right\|_2^2 .
\end{aligned}
\end{equation}
After the outer loop selects $\mathbf{q}^{\star}$, the evaluator combines it with the corresponding inner-loop solution $\mathbf{w}^{\star}(\mathbf{q}^{\star})$ to form the mixed fit $\theta_{\mathrm{mix}}^{\star}=(\mathbf{w}^{\star}(\mathbf{q}^{\star}),\mathbf{q}^{\star})$. This mixed fit is the primary result. STRIDE also runs a generic BFGS optimization~\citep{fletcher1987practical} as a fallback and retains the lower-NMSE solution as the final parameters $\boldsymbol{\theta}^{\star}$.

The fitted candidate is scored by an accuracy--complexity objective:
\begin{equation}
\begin{aligned}
\mathrm{Score}(f,\boldsymbol{\theta}^{\star};\mathcal{D})
= {} & W_{\mathrm{fit}}
\bigl[-\log(\mathrm{NMSE}+\epsilon)\bigr] \\
& - W_{\mathrm{comp}} C(f,\boldsymbol{\theta}^{\star}),
\end{aligned}
\label{eq:score}
\end{equation}
where $\mathrm{NMSE}$ measures predictive fidelity and $C(\cdot)$ penalizes structural and behavioral complexity, including effective parameter count, sensitivity, and curvature. The resulting parameters and score are then passed to critic--executor repair and semantic memory updating. This shared feedback lets later decisions depend on fitted behavior rather than the raw generated form. Additional details are provided in Appendix~\ref{app:ours_details} and \ref{sec:effect_mix}.

\subsection{Critic--Executor Agents: Reflective Repair}
\label{sec:method_refine}

For each evaluated candidate, STRIDE first applies a trigger policy $\mathrm{TriggerCritic}(s)$ to decide whether reflective repair should be activated. The policy selects promising but imperfect candidates according to score positivity and a trigger probability $\pi_c$, avoiding reflection budget on clearly poor candidates while retaining stochastic exploration among viable ones. Once triggered, two coordinated agents perform constrained local repair: a critic $\mathcal{M}_{\mathrm{critic}}$ and an executor $\mathcal{M}_{\mathrm{exec}}$. The critic examines the base equation $f$, fitted parameters, and score feedback, and proposes edits from the action space
\begin{equation}
\mathcal{A}=\{\textsc{Remove},\textsc{Simplify},\textsc{Add}\}.
\end{equation}
The executor then converts these edit instructions into executable refined candidates $\tilde{\mathcal{C}}$.

STRIDE reevaluates the refined candidates in two stages. A lightweight screening pass on a small data subset first retains only a few promising revisions. These finalists are then sent to the full evaluator for mixed fitting and scoring. The best refined candidate is compared with the original equation, and the better one is written back to semantic memory $\mathcal{B}$. Additional implementation details and a case study are provided in Appendix~\ref{sec:effect_critic}.

\subsection{Memory Module: Semantic Updating}
\label{sec:method_buffer}

After evaluation or reflection, STRIDE inserts the selected equation into a multi-island semantic memory $\mathcal{B}$ \citep{cranmer2023pysr, romeraparedes2024funsearch}. Before insertion, the equation is canonicalized, encoded with TF-IDF, and compared with existing cases so that structurally related hypotheses are assigned to the same semantic cluster $c$ \citep{shojaee2025llmsr}. Details of the TF-IDF construction are provided in Appendix~\ref{sec:tfidf}.

Within each cluster $c$, STRIDE retains only the highest-scoring equation as the elite representative. If a stronger equation enters the cluster, it replaces the current elite. In subsequent iterations, elite cases $\mathcal{E}$ are retrieved from the selected island and fed back into the generator agent $\mathcal{M}_{\mathrm{gen}}$ as in-context exemplars. This keeps memory compact and diverse while feeding evaluated discoveries back into later sampling.

\section{Experiments}
\label{sec:experiments}

In this section, we focus on the following questions:

\begin{tcolorbox}[
    enhanced,
    colback=black!2,
    colframe=black!15,
    boxrule=0.35pt,
    arc=1.5mm,
    sharp corners=southwest,
    sharp corners=southeast,
    left=1.2mm,right=1.2mm,top=1.2mm,bottom=1.2mm
]
\begin{enumerate}[leftmargin=1.2em, itemsep=2pt, topsep=2pt]
\item Does STRIDE improve equation discovery performance over strong symbolic regression and LLM-based baselines? (Section~\ref{sec:main_results})
\item Does STRIDE preserve more reliable structure under distribution shift? (Section~\ref{sec:main_results})
\item Which components are most responsible for STRIDE's gains, and what qualitative behaviors do they enable? (Section~\ref{sec:ablation_analysis})
\end{enumerate}
\end{tcolorbox}

\begin{table*}[t]
\centering
\caption{Main results on four symbolic regression benchmarks and four \textsc{LSR-Synth} benchmark suites. Acc\textcolor{red!70!black}{$\uparrow$} and NMSE\textcolor{green!50!black}{$\downarrow$} indicate that higher and lower values are better, respectively.
The best and second-best results are highlighted in \textcolor{red!50}{red} and \textcolor{blue!50}{blue}, respectively.}
\label{tab:main_results}
\scriptsize
\setlength{\tabcolsep}{1pt}
\renewcommand{\arraystretch}{1.02}
\resizebox{\textwidth}{!}{
\begin{tabular}{lcccccccccccc}
\toprule
\multirow{2}{*}{\textbf{Method}} 
& \multicolumn{3}{c}{\textbf{Oscillator 1}}
& \multicolumn{3}{c}{\textbf{Oscillator 2}}
& \multicolumn{3}{c}{\textbf{E. coli Growth}}
& \multicolumn{3}{c}{\textbf{Stress-Strain}} \\
\cmidrule(lr){2-4} \cmidrule(lr){5-7} \cmidrule(lr){8-10} \cmidrule(lr){11-13}
& {\fontsize{6.2}{6.6}\selectfont Acc$_{0.1}$\textcolor{red!70!black}{$\uparrow$}} & {\fontsize{6.2}{6.6}\selectfont Acc$_{0.001}$\textcolor{red!70!black}{$\uparrow$}} & {\fontsize{6.2}{6.6}\selectfont NMSE\textcolor{green!50!black}{$\downarrow$}}
& {\fontsize{6.2}{6.6}\selectfont Acc$_{0.1}$\textcolor{red!70!black}{$\uparrow$}} & {\fontsize{6.2}{6.6}\selectfont Acc$_{0.001}$\textcolor{red!70!black}{$\uparrow$}} & {\fontsize{6.2}{6.6}\selectfont NMSE\textcolor{green!50!black}{$\downarrow$}}
& {\fontsize{6.2}{6.6}\selectfont Acc$_{0.1}$\textcolor{red!70!black}{$\uparrow$}} & {\fontsize{6.2}{6.6}\selectfont Acc$_{0.001}$\textcolor{red!70!black}{$\uparrow$}} & {\fontsize{6.2}{6.6}\selectfont NMSE\textcolor{green!50!black}{$\downarrow$}}
& {\fontsize{6.2}{6.6}\selectfont Acc$_{0.1}$\textcolor{red!70!black}{$\uparrow$}} & {\fontsize{6.2}{6.6}\selectfont Acc$_{0.001}$\textcolor{red!70!black}{$\uparrow$}} & {\fontsize{6.2}{6.6}\selectfont NMSE\textcolor{green!50!black}{$\downarrow$}} \\
\midrule
\rowcolor{gray!10}\multicolumn{13}{l}{\textit{Baselines without LLMs}} \\
GPlearn & 10.41 & 0.12 & 1.83e{-}1 & 17.52 & 0.36 & 2.00e{-}1 & 0.92 & 0.00 & 1.00e{0} & 35.92 & 0.14 & 3.68e{-}1 \\
PySR & 97.98 & 10.29 & 4.71e{-}5 & 99.98 & 98.53 & 1.10e{-}9 & 10.52 & 0.08 & 1.02e{-}2 & 83.36 & 3.74 & 2.14e{-}2 \\
DSR & 11.19 & 0.09 & 9.74e{-}2 & 14.98 & 0.16 & 1.90e{-}1 & 3.52 & 0.00 & 6.89e{-}1 & 20.53 & 0.21 & 3.50e{-}1 \\
uDSR & 12.83 & 0.11 & 6.55e{-}2 & 16.67 & 0.15 & 1.96e{-}1 & 2.76 & 0.00 & 6.95e{-}1 & 7.14 & 0.00 & 3.37e{-}1 \\
\midrule
\rowcolor{gray!10}\multicolumn{13}{l}{\textit{GPT-5.1}} \\
LLM-SR & \cellcolor{blue!8}99.99 & \cellcolor{blue!8}95.64 & \cellcolor{blue!8}1.61e{-}7 & \cellcolor{blue!8}99.49 & \cellcolor{blue!8}43.16 & \cellcolor{blue!8}1.91e{-}6 & 1.24 & \cellcolor{blue!8}0.04 & 8.65e{-}1 & \cellcolor{blue!8}84.19 & \cellcolor{blue!8}2.43 & \cellcolor{blue!8}1.92e{-}2 \\
LaSR & 94.91 & 6.92 & 2.06e{-}4 & 32.59 & 0.27 & 9.69e{-}2 & \cellcolor{blue!8}2.28 & \cellcolor{blue!8}0.04 & \cellcolor{blue!8}1.21e{-}1 & 75.52 & \cellcolor{red!8}6.31 & 3.57e{-}2 \\
SR-LLM & 77.87 & 1.57 & 2.96e{-}3 & 17.48 & 0.52 & 2.00e{-}1 & 0.92 & 0.00 & 8.61e{-}1 & 62.83 & 0.69 & 7.91e{-}2 \\
DrSR & 99.94 & 48.44 & 1.54e{-}6 & 84.09 & 1.12 & 2.91e{-}3 & 1.08 & \cellcolor{blue!8}0.04 & 6.13e{-}1 & 77.18 & 0.69 & 2.74e{-}2 \\
STRIDE & \cellcolor{red!8}100.00 & \cellcolor{red!8}100.00 & \cellcolor{red!8}3.36e{-}18 & \cellcolor{red!8}100.00 & \cellcolor{red!8}100.00 & \cellcolor{red!8}1.12e{-}21 & \cellcolor{red!8}10.44 & \cellcolor{red!8}0.16 & \cellcolor{red!8}5.84e{-}3 & \cellcolor{red!8}88.28 & 1.25 & \cellcolor{red!8}1.82e{-}2 \\
\midrule
\rowcolor{gray!10}\multicolumn{13}{l}{\textit{Claude-3-Haiku}} \\
LLM-SR 
& \cellcolor{blue!8}99.77 & \cellcolor{blue!8}38.89 & 2.11e{-}5 & \cellcolor{blue!8}99.99 & \cellcolor{blue!8}99.52 & \cellcolor{blue!8}6.01e{-}9 & \cellcolor{blue!8}11.64 & \cellcolor{red!8}0.20 & \cellcolor{blue!8}1.47e{-}2 & 77.05 & 1.32 & 2.62e{-}2 \\
LaSR 
& 96.57 & 5.37 & 7.14e{-}5 & 94.77 & 2.94 & 1.41e{-}4 & 2.92 & 0.00 & 1.38e{-}1 & \cellcolor{blue!8}80.24 & \cellcolor{red!8}2.57 & \cellcolor{blue!8}2.05e{-}2 \\
SR-LLM 
& 67.23 & 0.77 & 7.61e{-}3 & 17.48 & 0.52 & 2.00e{-}1 & 1.24 & 0.00 & 3.24e{-}1 & 73.86 & 1.73 & 4.84e{-}2 \\
DrSR 
& 99.56 & 38.76 & \cellcolor{blue!8}1.90e{-}6 & 75.89 & 0.75 & 3.80e{-}2 & 1.04 & 0.00 & 7.22e{-}1 & 66.92 & 0.49 & 7.16e{-}2 \\
\textbf{STRIDE}
& \cellcolor{red!8}100.00 & \cellcolor{red!8}100.00 & \cellcolor{red!8}2.52e{-}15 
& \cellcolor{red!8}100.00 & \cellcolor{red!8}99.99 & \cellcolor{red!8}6.79e{-}12 
& \cellcolor{red!8}14.60 & \cellcolor{red!8}0.20 & \cellcolor{red!8}3.79e{-}3 
& \cellcolor{red!8}85.64 & \cellcolor{blue!8}2.43 & \cellcolor{red!8}1.55e{-}2 \\
\midrule
\midrule
\multirow{2}{*}{\textbf{Method}} 
& \multicolumn{3}{c}{\textbf{CRK}} 
& \multicolumn{3}{c}{\textbf{PO}} 
& \multicolumn{3}{c}{\textbf{MatSci}} 
& \multicolumn{3}{c}{\textbf{BPG}} \\
\cmidrule(lr){2-4} \cmidrule(lr){5-7} \cmidrule(lr){8-10} \cmidrule(lr){11-13}
& {\fontsize{6.0}{6.4}\selectfont Acc$_{0.1}$\textcolor{red!70!black}{$\uparrow$}} & {\fontsize{6.0}{6.4}\selectfont Acc$_{\max,0.1}$\textcolor{red!70!black}{$\uparrow$}} & {\fontsize{6.0}{6.4}\selectfont NMSE\textcolor{green!50!black}{$\downarrow$}}
& {\fontsize{6.0}{6.4}\selectfont Acc$_{0.1}$\textcolor{red!70!black}{$\uparrow$}} & {\fontsize{6.0}{6.4}\selectfont Acc$_{\max,0.1}$\textcolor{red!70!black}{$\uparrow$}} & {\fontsize{6.0}{6.4}\selectfont NMSE\textcolor{green!50!black}{$\downarrow$}}
& {\fontsize{6.0}{6.4}\selectfont Acc$_{0.1}$\textcolor{red!70!black}{$\uparrow$}} & {\fontsize{6.0}{6.4}\selectfont Acc$_{\max,0.1}$\textcolor{red!70!black}{$\uparrow$}} & {\fontsize{6.0}{6.4}\selectfont NMSE\textcolor{green!50!black}{$\downarrow$}}
& {\fontsize{6.0}{6.4}\selectfont Acc$_{0.1}$\textcolor{red!70!black}{$\uparrow$}} & {\fontsize{6.0}{6.4}\selectfont Acc$_{\max,0.1}$\textcolor{red!70!black}{$\uparrow$}} & {\fontsize{6.0}{6.4}\selectfont NMSE\textcolor{green!50!black}{$\downarrow$}} \\
\midrule
\rowcolor{gray!10}\multicolumn{13}{l}{\textit{GPT-5.1}} \\
LLM-SR       
& \cellcolor{blue!8}90.44 & \cellcolor{blue!8}60.00 & \cellcolor{blue!8}7.06e{-}5 
& \cellcolor{blue!8}75.48 & 0.00 & \cellcolor{blue!8}7.75e{-}4 
& \cellcolor{blue!8}99.00 & \cellcolor{blue!8}60.00 & \cellcolor{red!8}8.59e{-}4 
& \cellcolor{blue!8}81.00 & \cellcolor{blue!8}60.00 & \cellcolor{blue!8}3.53e{-}5 \\

SR-Scientist 
& 83.56 & 40.00 & 7.76e{-}5 
& 55.44 & 0.00 & 6.36e{-}3 
& \cellcolor{red!8}100.00 & \cellcolor{red!8}100.00 & \cellcolor{blue!8}2.14e{-}3 
& 68.08 & 20.00 & 2.71e{-}4 \\

\textbf{STRIDE}
& \cellcolor{red!8}100.00 & \cellcolor{red!8}100.00 & \cellcolor{red!8}1.83e{-}9 
& \cellcolor{red!8}75.68 & 0.00 & \cellcolor{red!8}4.17e{-}4 
& \cellcolor{red!8}100.00 & \cellcolor{red!8}100.00 & \cellcolor{blue!8}2.14e{-}3 
& \cellcolor{red!8}91.88 & \cellcolor{red!8}80.00 & \cellcolor{red!8}4.16e{-}7 \\
\midrule
\rowcolor{gray!10}\multicolumn{13}{l}{\textit{Claude-3-Haiku}} \\
LLM-SR       
& \cellcolor{blue!8}91.96 & 40.00 & \cellcolor{blue!8}3.31e{-}5 
& 65.96 & 0.00 & \cellcolor{red!8}2.02e{-}4 
& \cellcolor{red!8}100.00 & \cellcolor{red!8}100.00 & \cellcolor{red!8}9.68e{-}5 
& 74.60 & 0.00 & 6.31e{-}4 \\

SR-Scientist 
& \cellcolor{blue!8}92.56 & \cellcolor{blue!8}80.00 & 4.75e{-}5 
& \cellcolor{red!8}78.04 & 0.00 & \cellcolor{blue!8}5.28e{-}4 
& \cellcolor{red!8}100.00 & \cellcolor{red!8}100.00 & \cellcolor{blue!8}2.13e{-}3 
& \cellcolor{blue!8}76.92 & \cellcolor{blue!8}20.00 & \cellcolor{blue!8}8.17e{-}5 \\

\textbf{STRIDE}
& \cellcolor{red!8}100.00 & \cellcolor{red!8}100.00 & \cellcolor{red!8}5.35e{-}9 
& \cellcolor{blue!8}77.88 & 0.00 & 1.72e{-}3 
& \cellcolor{red!8}100.00 & \cellcolor{red!8}100.00 & 2.18e{-}3 
& \cellcolor{red!8}99.92 & \cellcolor{red!8}80.00 & \cellcolor{red!8}4.82e{-}9 \\

\bottomrule
\end{tabular}
}
\end{table*}

\subsection{Setup}
\label{sec:exp_setup}

\paragraph{Benchmarks.}
We evaluate STRIDE on two benchmark groups: four representative \textsc{LLM-SR} tasks~\citep{shojaee2025llmsr} and four \textsc{LSR-Synth} suites from \textsc{LLM-SRBench}~\citep{shojaee2025llmsrbench}; detailed equations are listed in Appendix~\ref{app:dataset}. For \textsc{LSR-Synth}, we use the top five tasks in each suite according to the difficulty ranking. All benchmarks are evaluated under both in-domain (ID) and out-of-domain (OOD) settings.

\paragraph{Baselines.}
We compare against traditional symbolic regression baselines, including \textsc{gplearn}~\citep{stephens2022gplearn}, \textsc{PySR}~\citep{cranmer2023pysr}, \textsc{DSR}~\citep{petersen2021dsr}, and \textsc{uDSR}~\citep{landajuela2022udsr}, as well as recent LLM-based methods: \textsc{LLM-SR}~\citep{shojaee2025llmsr}, \textsc{LaSR}~\citep{grayeli2024lasr}, \textsc{SR-LLM}~\citep{guo2025srllm}, \textsc{SR-Scientist}~\citep{xia2026srscientist}, and \textsc{DrSR}~\citep{wang2025drsr}. Additional baseline details are deferred to Appendix~\ref{app:baseline_details}.

\paragraph{Metrics.}
We report both fitting error and tolerance-based accuracy. Specifically, we use NMSE to measure predictive fidelity while normalizing for target scale, together with pointwise tolerance accuracy $\mathrm{ACC}@\tau$ and the stricter split-level accuracy $\mathrm{ACC}_{\max}@\tau$:
\begin{equation}
\begin{aligned}
\mathrm{NMSE}(\hat{\mathbf{y}}, \mathbf{y})
=
\frac{\frac{1}{N}\sum_{i=1}^{N}(\hat{y}_i-y_i)^2}
{\mathrm{Var}(\mathbf{y})}, \\
\mathrm{ACC}@\tau
=
\frac{1}{N}\sum_{i=1}^{N}
\mathbb{I}
\left(
\frac{|\hat{y}_i-y_i|}{|y_i|+\epsilon}
\le \tau
\right), \\
\mathrm{ACC}_{\max}@\tau
=
\mathbb{I}
\left(
\max_{1 \le i \le N}
\frac{|\hat{y}_i-y_i|}{|y_i|+\epsilon}
\le \tau
\right).
\end{aligned}
\end{equation}

\paragraph{Configuration.}
We align the search budget across LLM-based methods and evaluate \textsc{GPT-5.1}~\citep{openai2025gpt51} and \textsc{Claude-3-Haiku}~\citep{anthropic2024claude3haiku} with the same prompt template and decoding temperature. STRIDE generates at most 2000 candidate expressions per dataset, and \textsc{GPT-5.1} is used as the default backbone unless otherwise specified. Full hyperparameter settings are provided in Appendix~\ref{app:baseline_details} and Appendix~\ref{app:ours_details}.

\begin{figure*}[t]
    \centering
    \includegraphics[width=\textwidth]{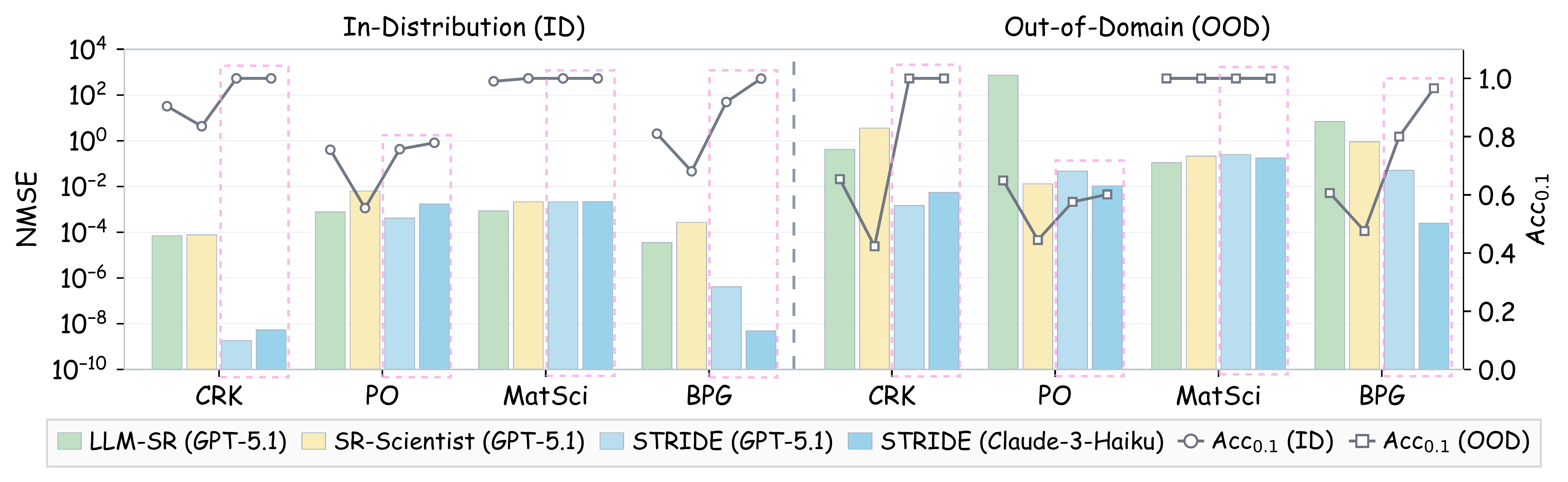}
    \caption{ID and OOD performance on the LSR-Synth suites. Bars report NMSE\textcolor{green!50!black}{$\downarrow$} and line markers report Acc$_{0.1}$\textcolor{red!70!black}{$\uparrow$} for baselines and STRIDE. Complete OOD results for all benchmarks and LLMs are provided in Appendix~\ref{app:ood_results}.}
    \label{fig:ood_performance}
    \end{figure*}

\begin{figure*}[t]
    \centering
    \includegraphics[width=\linewidth]{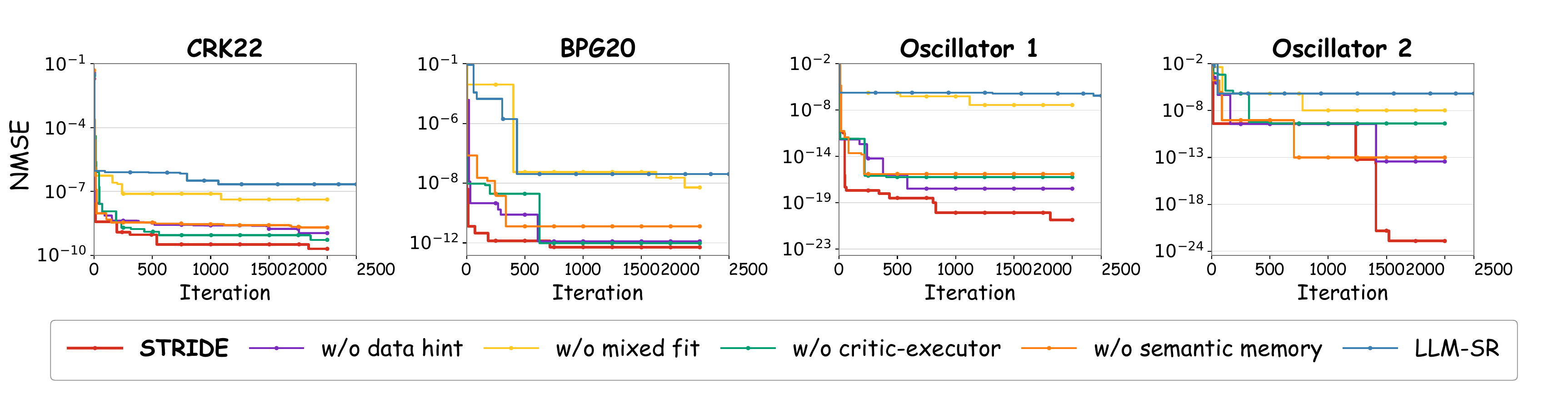}
    \caption{Iteration analysis of STRIDE and its ablated variants. The full system converges more consistently than variants with individual modules removed.}
    \label{fig:ablation_iterations}
\end{figure*}

\subsection{Main Results}
\label{sec:main_results}

\paragraph{STRIDE Discovers More Reliable Equations Than Strong Baselines.}
Table~\ref{tab:main_results} answers the first question affirmatively: STRIDE consistently improves equation discovery over strong symbolic regression and LLM-based baselines. Relative to LLM-SR, STRIDE improves average Acc$_{0.1}$ by 3.45\% and 2.95\% on the four main benchmarks under GPT-5.1 and Claude-3-Haiku, respectively; on LSR-Synth, the gains further increase to 5.16\% and 11.32\%. The stricter Acc$_{\max,0.1}$ results reinforce this conclusion: STRIDE achieves the best or tied-best score in nearly all LSR-Synth domains under both backbones, indicating that its accuracy does not come from fitting only a subset of points but from discovering equations that remain closer to the ground-truth structure as a whole.

\paragraph{STRIDE Preserves More Reliable Structure Under Distribution Shift.}
Figure~\ref{fig:ood_performance} answers the second question by showing that STRIDE achieves the highest accuracy across all domains under distribution shift, with consistently stronger OOD Acc$_{0.1}$ and lower or competitive NMSE. This indicates that STRIDE is not merely fitting the training region more closely; it is more often recovering equations whose structure transfers to unseen regimes, reducing the risk that the improvement comes from overfitting. The effect is especially clear on \textit{CRK} and \textit{BPG}, where STRIDE maintains strong OOD accuracy while competing methods drift after leaving the observed region.

\subsection{Ablation Analysis}
\label{sec:ablation_analysis}

\paragraph{Each STRIDE Component Contributes to Reliability.}
Table~\ref{tab:ablation_ood} and Figure~\ref{fig:ablation_iterations} show that each module contributes to STRIDE's final performance and search dynamics. Removing data hints, mixed fitting, the critic--executor, or semantic memory degrades at least one key metric on the OOD oscillator tasks, with mixed fitting producing the largest drop in NMSE and strict accuracy. Even with individual modules removed, STRIDE variants generally remain stronger than LLM-SR. This confirms that reliables parameter feedback is central, but not sufficient by itself. The iteration curves further show that the full system converges more consistently than its ablated variants. Overall, the modules contribute complementary gains to the closed-loop search.
\begin{table}[!t]
\centering
\caption{Ablation results on the OOD oscillator tasks. ``w/o data hints'' uses only the problem description and exemplar equations during sampling, without data-statistical hints; ``w/o mixed fit'' uses only BFGS for parameter fitting in the evaluation stage; ``w/o critic--executor'' removes the reflection stage; and ``w/o semantic memory'' replaces semantic clustering with score-based experience management.}
\label{tab:ablation_ood}
\scriptsize
\setlength{\tabcolsep}{3pt}
\renewcommand{\arraystretch}{1.05}
\resizebox{\linewidth}{!}{
\begin{tabular}{lccc}
\toprule
\textbf{Method} & \textbf{NMSE}\textcolor{green!50!black}{$\downarrow$} & \textbf{Acc$_{0.1}$}\textcolor{red!70!black}{$\uparrow$} & \textbf{Acc$_{0.001}$}\textcolor{red!70!black}{$\uparrow$} \\
\midrule
\textbf{Ours (Oscillator 1)} & \textbf{5.97e{-}12} & \textbf{\underline{100.00}} & \textbf{\underline{100.00}} \\
\quad w/o data hints & 5.64e{-}10 & \underline{100.00} & 99.99 \\
\quad w/o mixed fit & 4.66e{-}6 & \underline{100.00} & 69.61 \\
\quad w/o critic--executor & 4.92e{-}8 & \underline{100.00} & 89.75 \\
\quad w/o semantic memory & 9.47e{-}8 & \underline{100.00} & 84.87 \\
\quad LLM-SR (none) & 2.85e{-}4 & 99.97 & 41.21 \\
\midrule
\textbf{Ours (Oscillator 2)} & \textbf{4.48e{-}21} & \textbf{\underline{100.00}} & \textbf{\underline{100.00}} \\
\quad w/o data hints & 4.76e{-}9 & \underline{100.00} & 99.99 \\
\quad w/o mixed fit & 1.85e{-}4 & 96.01 & 58.67 \\
\quad w/o critic--executor  & 1.40e{-}6 & \underline{100.00} & 94.17 \\
\quad w/o semantic memory & 4.61e{-}9 & \underline{100.00} & \underline{100.00} \\
\quad LLM-SR (none) & 3.05e{-}3 & 86.28 & 19.33 \\
\bottomrule
\end{tabular}
}
\end{table}

\begin{figure}[t]
\centering
\includegraphics[width=.85\linewidth]{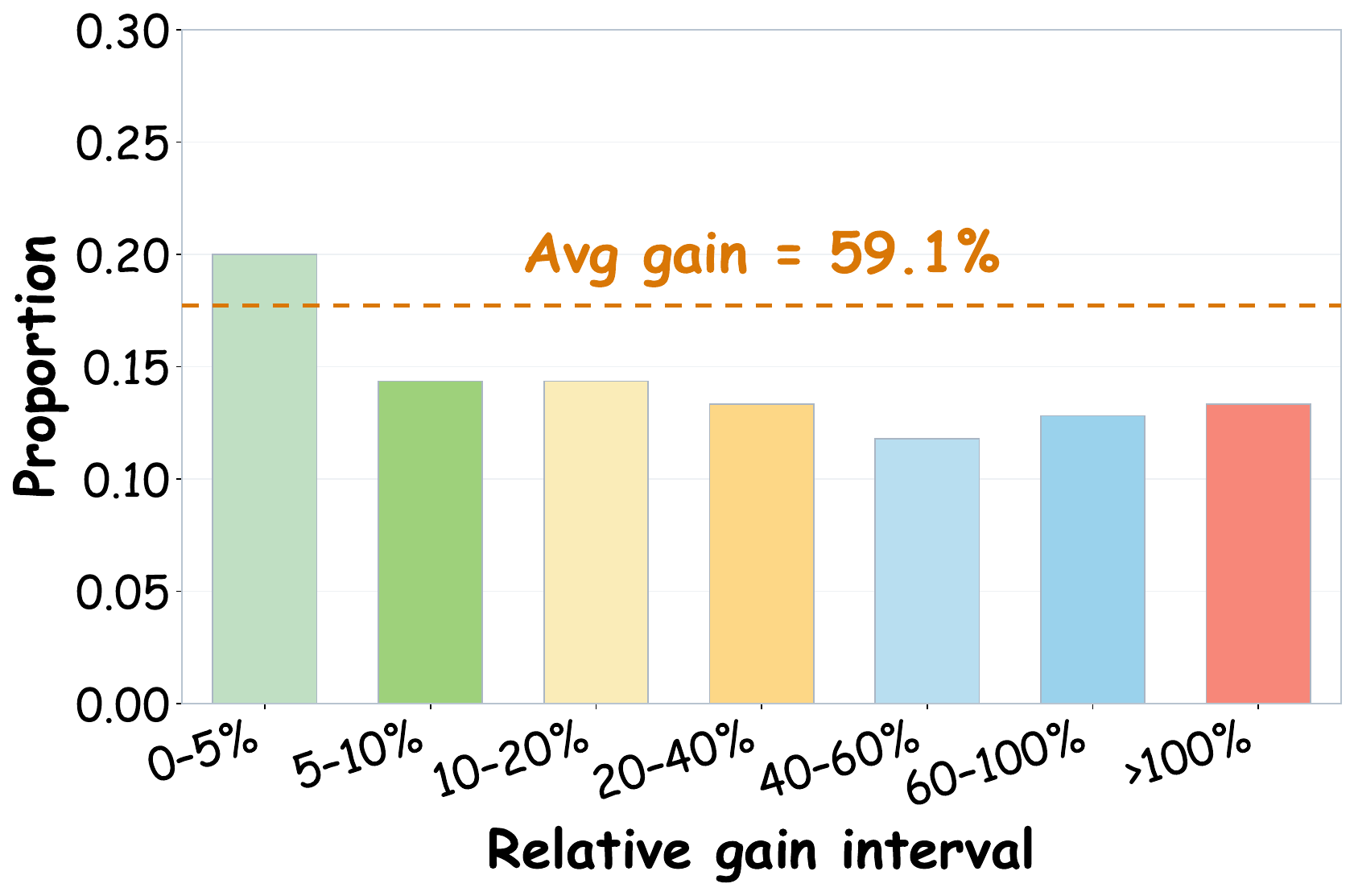}
\caption{Reflection gains from critic--executor repair. Reflection often improves promising candidates, showing the benefit of targeted local revision.}
\label{fig:critic_improve}
\end{figure}

\begin{figure}[t]
    \centering
    \includegraphics[width=.95\linewidth]{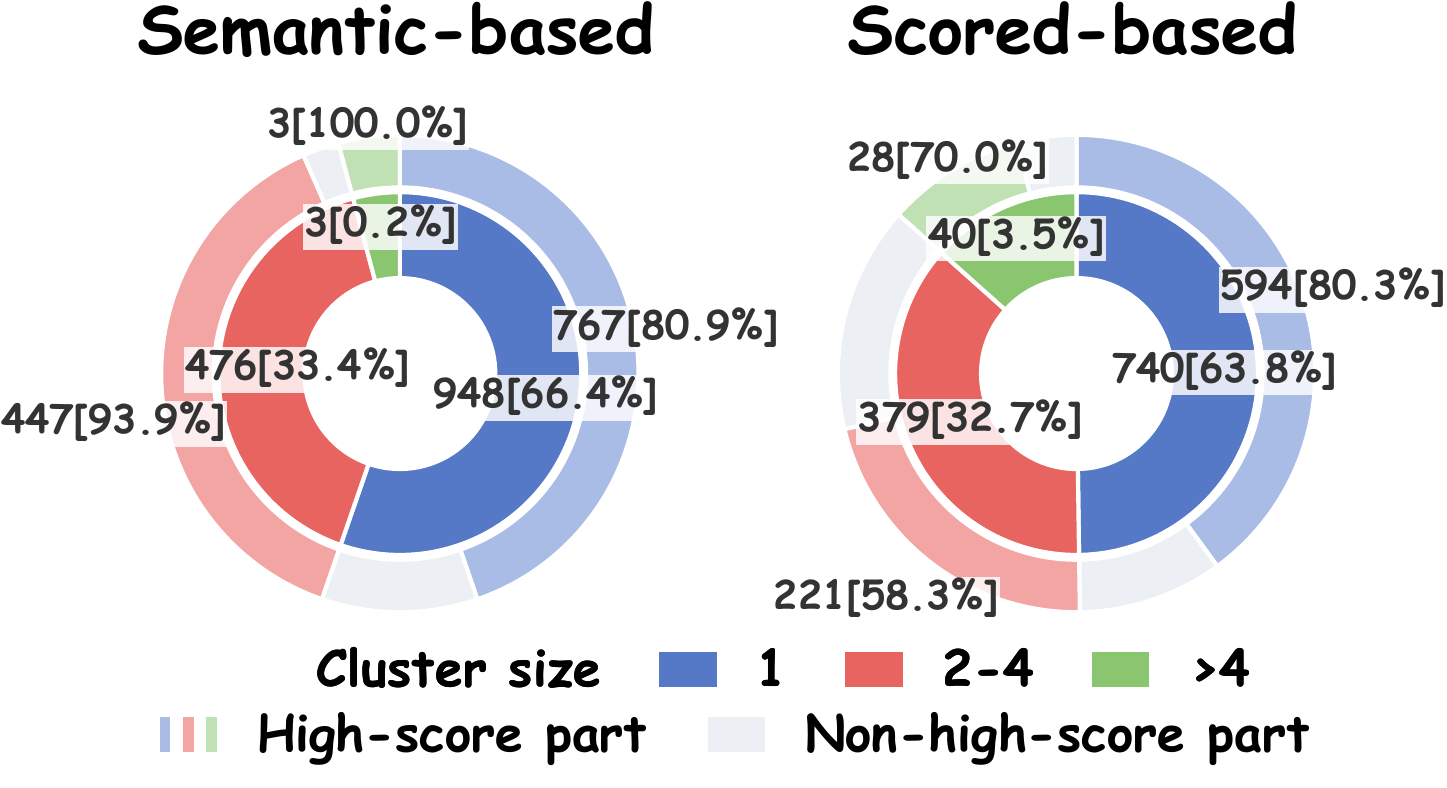}
    \caption{High-score semantic versus score-based memory grouping on \textit{Oscillator 1}. The inner ring shows the distribution of all clusters by size, and the outer ring shows the number and proportion of clusters in each size group that contain high-score equations. High-score candidates are defined as Acc$_{0.1}>0.9$.}
    \label{fig:cluster_compare}
\end{figure}

\begin{table}[t]
\centering
\caption{Estimated per-problem cost and relative OOD performance of LLM-based methods under our evaluation setting. Costs are approximate; relative metrics are normalized to the lowest method value.}
\label{tab:cost_compare_main}
\scriptsize
\setlength{\tabcolsep}{5pt}
\renewcommand{\arraystretch}{1.05}
\resizebox{\linewidth}{!}{
\begin{tabular}{lcccc}
\toprule
\textbf{Method} & \textbf{Tokens} & \textbf{Time} & \textbf{Rel. Acc$_{0.1}$ \textcolor{red!70!black}{$\uparrow$}} & \textbf{Rel. NMSE \textcolor{green!50!black}{$\downarrow$}} \\
\midrule
LLM-SR & $\sim$2.6M & $\sim$6.0h & 2.49$\times$ & 7$\times$ \\
LaSR & $\sim$1.3M & $\sim$2.8h & 1.68$\times$ & 76$\times$ \\
SR-LLM & $\sim$0.7M & $\sim$5.5h & 1.00$\times$ & 6489$\times$ \\
DrSR & $\sim$3.8M & $\sim$7.2h & 1.86$\times$ & 118$\times$ \\
\rowcolor{yellow!12}\textbf{STRIDE} & $\sim$3.5M & $\sim$6.7h & \textbf{3.26$\times$} & \textbf{1$\times$} \\
\bottomrule
\end{tabular}
}
\end{table}

\subsection{Further Analysis}
\label{sec:further_analysis}

\begin{figure*}[t]
    \centering
    \includegraphics[width=.96\textwidth]{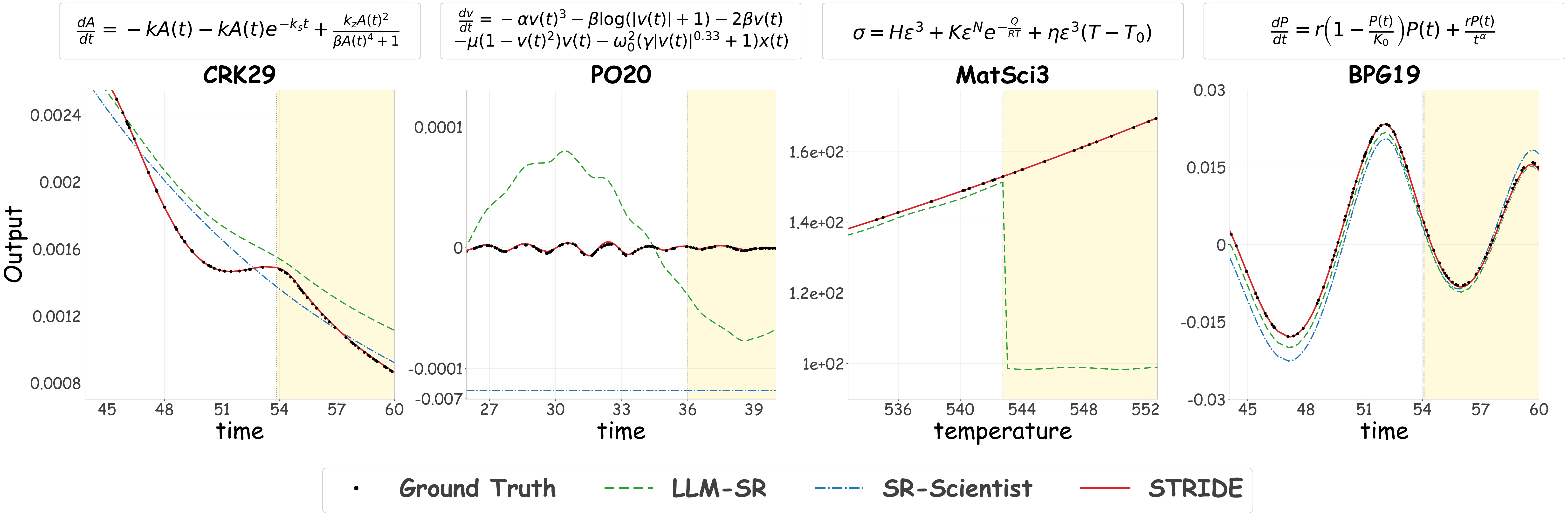}
    \caption{Cross-domain LSR-Synth case analysis. Shaded regions denote OOD intervals. STRIDE more closely tracks the ground truth across CRK19, PO20, MatSci3, and BPG19, indicating more reliable structural recovery.}
    \label{fig:case_analysis}
    \vspace{-0.2cm}
\end{figure*}

\paragraph{Reflective Repair Converts Feedback Into Local Improvements.}
Figure~\ref{fig:critic_improve} shows that critic--executor repair improves many reflected candidates, indicating that reflection acts as targeted repair rather than additional unconstrained generation. This comes from conditioning the critic on fitted parameters, score feedback, and task context, so that edits are grounded in the evaluated behavior of the current equation. Ablations in Table~\ref{tab:ablation_ood} provide supporting evidence: removing the critic--executor reduces final precision on the oscillator tasks. Overall, STRIDE turns evaluator feedback into local symbolic improvements.

\paragraph{Semantic Memory Helps Recover Clear Complex Structures.}
STRIDE is especially useful when equations contain clear but nontrivial symbolic mechanisms, such as the oscillatory, \textit{CRK}, and \textit{MatSci} cases studied here. Figure~\ref{fig:cluster_compare} provides a memory-level explanation: the inner ring summarizes memory clusters by size, while the outer ring marks how many clusters in each group contain high-score equations. Compared with score-based retention, semantic memory keeps high-score equations in more singleton and small clusters, indicating that useful hypotheses are structurally diverse rather than concentrated in redundant groups. This diversity gives later prompts access to distinct high-quality mechanisms, improving search over clear and interpretable equation structures.

\paragraph{STRIDE Balances Cost and Search Efficiency.}
STRIDE introduces extra evaluator and reflection calls, but uses them to make search more productive. Figure~\ref{fig:ablation_iterations} shows that feedback from fitting, repair, and memory helps the system reach strong candidates early, while Figure~\ref{fig:mix_compare} shows that mixed fitting improves parameter estimation with manageable overhead. As summarized in Table~\ref{tab:cost_compare_main}, STRIDE achieves substantially stronger OOD performance without requiring markedly higher token or runtime budgets than comparable LLM-based baselines.

\paragraph{Case Study.}
Figure~\ref{fig:case_analysis} connects memory-level diversity to equation-level behavior. Across CRK19, PO20, MatSci3, and BPG19, STRIDE more closely follows the ground truth in OOD regions, suggesting that it preserves useful structural alternatives rather than only improving local fit. Even for coupled nonlinear components such as $x\exp(x)$, STRIDE better retains the dominant mechanisms that control extrapolation, yielding smoother transitions and smaller post-shift drift. Additional examples are provided in Appendix~\ref{app:case_study}.

\section{Related Work}

\paragraph{Symbolic Regression and LLM-based Equation Discovery.}
Symbolic regression seeks interpretable equations that explain observed data. Classical methods search over expression trees with genetic programming, multi-island evolution, decomposition, or physics-informed structure search \citep{koza1992genetic,cranmer2023pysr,udrescu2020aifeynman,sun2023symbolicphysicslearner}. These systems remain strong foundations, but their success still depends on balancing symbolic structure search with reliable parameter fitting, since poor coefficient estimation can make useful structures appear unpromising \citep{more1978lm,nocedal2006numerical,negishi2024twostage,zeng2025twostage}. Recent LLM-based methods use language models as priors over symbolic form, executable programs, and scientific patterns. Some frame discovery as in-context function induction, automatic equation/model discovery, or library-conditioned program search, where generated candidates are fitted and stored for later prompts \citep{merler2024icsr,du2024llm4edlargelanguagemodels,li2024automatedstatisticalmodeldiscovery,shojaee2025llmsr}. Others add concept abstraction, retrieval, data-aware reasoning, or physics-guided agent steps to improve candidate proposal and search direction \citep{grayeli2024lasr,guo2025srllm,wang2025drsr,xia2026srscientist,yang2026thinklikescientistphysicsguided}. Our work follows this LLM-guided line, but emphasizes fitting-aware feedback, targeted repair, and diversity-preserving memory as coordinated parts of the discovery loop.

\paragraph{Agentic Scientific Discovery.}
LLM agents have increasingly been used to coordinate tools, memory, and reflection for scientific reasoning and discovery. Tool-augmented, autonomous, and survey work covers scientific problem solving, symbolic solver use, self-improvement, chemistry tool use, and broader research workflows \citep{ma-etal-2024-sciagent,pan-etal-2023-logic,huang-etal-2023-large,bran2023chemcrowaugmentinglargelanguagemodels,lu2024aiscientistfullyautomated,schmidgall2025agentlaboratoryusingllm,zheng2025automationautonomysurveylarge}. Related agent studies also explore reflection learning, critic guidance, adaptive search, automated paper auditing, and reusable skill memories for long-horizon tasks \citep{gupta-etal-2024-metareflection,xiang-etal-2024-retrospex,sun2026autosearchadaptivesearchdepth,tu2026paperauditbenchbenchmarkingerrordetection,tu2026dynamicdualgranularityskillbank}. These works motivate decomposing complex tasks into tool use, search, critique, and memory; STRIDE grounds this coordination in fitted symbolic candidates for equation discovery.

\section{Conclusion}

We present \textbf{STRIDE}, a self-reflective agent framework for reliable LLM-based equation discovery. STRIDE coordinates data-aware generation, mixed-fitting feedback, critic--executor repair, and semantic memory to move beyond generation-only search. Experiments across multi-domain benchmarks and LLM backbones show consistent gains in accuracy, OOD robustness, and structural recovery, highlighting the value of feedback-driven agent coordination for automatic scientific modeling.

\clearpage

\section*{Limitations}

We summarize key practical limitations:
\begin{itemize}[leftmargin=*, itemsep=0pt]

\item \textbf{Parameter-role identification.}
STRIDE uses heuristic AST-based analysis to separate linear and nonlinear parameter roles. This may be less stable for deeply coupled expressions, although the BFGS fallback helps maintain robustness.

\item \textbf{Semantic equivalence.}
Semantic memory relies on canonicalization and surface-form similarity. Equivalent equations written in different algebraic forms may therefore lead to imperfect clustering or redundant entries.

\item \textbf{Real-world deployment.}
Our experiments use benchmark datasets with controlled ID/OOD splits. Applying STRIDE to noisier real-world scientific data may require additional uncertainty estimation and domain validation.

\end{itemize}



\section*{Ethics Statement}
This work studies automatic equation discovery on benchmark scientific datasets and does not involve human subjects, private data, or personally identifiable information. We do not anticipate direct negative societal impacts. Discovered equations should nevertheless be independently validated before use in scientific or engineering decision-making.

\bibliography{stride}

\clearpage

\appendix

\section{Details of Our Inference Framework}
\label{app:ours_details}
\label{sec:tfidf}

Algorithm~\ref{alg:stride} summarizes the full inference loop of STRIDE, including data-aware sampling, mixed-fitting-based evaluation, critic--executor reflection, and semantic-memory updating.

We run our framework with a maximum budget of 2000 candidate equations per dataset. 
At each iteration, STRIDE constructs the in-context experience examples from a 10-island semantic memory buffer. 
Following the multi-island sampling strategy used in LLM-guided program search~\citep{romeraparedes2024funsearch, shojaee2025llmsr}, we first randomly select one island from the available islands. 
Within the selected island, equations are organized into semantic clusters maintained during memory updating. 
The highest-scoring equation in each cluster is regarded as the elite case, and its score is used as the cluster signature. 
Clusters with higher scores are sampled with larger probability. 
Let $s_i$ denote the signature score of cluster $c_i$. 
The probability of choosing cluster $c_i$ is
\begin{equation}
P_i =
\frac{\exp(s_i / \tau_c)}
{\sum_{j}\exp(s_j / \tau_c)} ,
\end{equation}
where $\tau_c=0.1$ is the cluster-sampling temperature. 
After a cluster is selected, STRIDE directly uses the highest-scoring equation in that cluster as the retrieved exemplar, rather than further preferring shorter programs inside the cluster. 
In this way, each prompt receives 2 elite exemplars and generates 4 candidate equation skeletons. 
When enabled, data hints extracted from the training split are injected every 25 epochs. 

\begin{algorithm}[t]
    \caption{STRIDE}
    \label{alg:stride}
    \footnotesize
    \begin{algorithmic}[1]
    \Statex \textbf{Input:} LLM $\mathcal{M}$, dataset $\mathcal{D}$, task specification $\mathcal{T}$, iterations $N$, semantic memory buffer $\mathcal{B}$
    \Statex \textbf{Output:} best equation $f^*$, best score $s^*$
    
    \State $h \gets \algblue{\mathrm{BuildDataHint}}(\mathcal{D})$
    \State $f^{*}, s^{*} \gets \varnothing, -\infty$
    
    \For{$i = 1$ to $N$}
        \Statex \algphase{A. Sampling}
        \State Sample elite cases $\mathcal{E}$ from memory $\mathcal{B}$
        \State $p \gets \algblue{\mathrm{GeneratePrompt}}(\mathcal{E},\mathcal{T},h)$
        \State $\mathcal{C} \gets \{f_j\}_{j=1}^{b},\; f_j \sim \mathcal{M}_{\mathrm{gen}}(\cdot \mid p)$
        \For{$f \in \mathcal{C}$}
            \Statex \algphase{B. Evaluation}
            \State $\theta^{*} \gets \algblue{\mathrm{MixedOptimize}}(f,\mathcal{D})$
            \State $s \gets \mathrm{Score}(f,\theta^{*};\mathcal{D})$
            \Statex \algphase{C. Reflection}
            \If{$\mathrm{TriggerCritic}(s)$}
                \State $critic \gets \algblue{\mathcal{M}_{\mathrm{critic}}}(\cdot \mid f,\theta^{*},\mathcal{D})$
                \State $\tilde{\mathcal{C}} \gets \algblue{\mathcal{M}_{\mathrm{exec}}}(\cdot \mid f,critic)$
                \For{$\tilde f \in \tilde{\mathcal{C}}$}
                    \State $\tilde{\theta} \gets \algblue{\mathrm{MixedOptimize}}(\tilde f,\mathcal{D})$
                    \State $\tilde s \gets \mathrm{Score}(\tilde f,\tilde{\theta};\mathcal{D})$
                \EndFor
                \State $(f,s) \gets \mathrm{BestOf}((f,s),\{(\tilde f,\tilde s)\})$
            \EndIf
            \Statex \algphase{D. Updating}
            \State $c \gets$ cluster ID of $f$ in $\mathcal{B}$ by \algblue{TF-IDF} similarity; see Eqs.~\eqref{equation_tfidf} and~\eqref{equation_sim}.
            \State $\mathcal{B} \gets \algblue{\mathrm{SemanticInsert}}(\mathcal{B},c,(f,s))$
            \If{$s > s^{*}$}
                \State $f^{*}, s^{*} \gets f, s$
            \EndIf
        \EndFor
    \EndFor
    
    \State \Return $f^{*}, s^{*}$
    \end{algorithmic}
\end{algorithm}

After the generator proposes candidate skeletons, STRIDE enters the evaluation stage, where the mixed-fitting evaluator first fits free parameters and then computes score feedback for candidate assessment. To improve parameter identification and avoid underestimating promising skeletons, we adopt a mixed fitting strategy. 
AST analysis classifies parameters as nonlinear when they appear inside nonlinear functions, as exponents, or in products with other parameter-dependent terms; the remaining active parameters are treated as linear. Given a nonlinear parameter vector $\mathbf{q}$, we compute a bias term $\mathbf{b}(\mathbf{q})$ by setting all linear coefficients to zero. For each linear coefficient $\mathbf{w}_k$, we then construct a unit probe
\begin{equation}
\boldsymbol{\phi}_k(\mathbf{q})
=
f(\mathbf{u}; \mathbf{w}_k = 1,\ \mathbf{w}_{j \neq k}=0,\ \mathbf{q}) - \mathbf{b}(\mathbf{q}).
\end{equation}
Stacking these probes yields the design matrix $\Phi(\mathbf{q})$. The conditional linear coefficients are solved by ridge least squares with coefficient $10^{-10}$:
\begin{equation}
\begin{aligned}
\mathbf{w}^{\star}(\mathbf{q})
= \arg\min_{\mathbf{w}}\;&
\left\|
\Phi(\mathbf{q}) \mathbf{w}
- \big(\mathbf{y}-\mathbf{b}(\mathbf{q})\big)
\right\|_2^2 \\
&+ \lambda \|\mathbf{w}\|_2^2.
\end{aligned}
\end{equation}
The nonlinear parameters are optimized by 4-start Powell search over the reduced objective
\begin{equation}
J(\mathbf{q})
=
\frac{1}{\mathrm{Var}(\mathbf{y})}
\cdot
\frac{1}{N}
\left\|
f\big(\mathbf{u}; \mathbf{w}^{\star}(\mathbf{q}), \mathbf{q}\big) - \mathbf{y}
\right\|_2^2.
\end{equation}
The resulting mixed solution is compared against a multi-start BFGS fallback, and the lower-NMSE solution is retained as the fitted candidate.

Given the fitted parameters returned by the evaluator, STRIDE computes the score feedback using $W_{\mathrm{fit}}=0.7$, $W_{\mathrm{comp}}=0.3$, and $\epsilon=10^{-12}$. The complexity term is defined as
\begin{equation}
\begin{aligned}
C = {} & K_{\mathrm{param}}n_{\mathrm{eff}}
+K_{\mathrm{sens}}C_{\mathrm{sens}} \\
& +K_{\mathrm{curv}}\log(1+C_{\mathrm{curv}}),
\end{aligned}
\end{equation}
where $n_{\mathrm{eff}}$ is the number of numerically active parameters, and $C_{\mathrm{sens}}$ and $C_{\mathrm{curv}}$ penalize excessive sensitivity and curvature. We set $K_{\mathrm{param}}=1.0$, $K_{\mathrm{sens}}=0.05$, and $K_{\mathrm{curv}}=0.01$. The effective parameter count is estimated by perturbing each parameter and checking whether the output changes above a small relative threshold. Sensitivity and curvature are computed on at most 300 training points using first- and second-order finite differences.

The critic--executor reflection stage is activated by $\mathrm{TriggerCritic}(s)$, which triggers reflection when the score feedback is above the threshold $0$ and a Bernoulli trial with probability $\pi_c=0.4$ succeeds. For each selected base candidate, the critic agent $\mathcal{M}_{\mathrm{critic}}$ receives the candidate code, fitted parameters, score feedback, input variables, and task description, and proposes constrained edits from the action space $\{\textsc{Remove},\textsc{Simplify},\textsc{Add}\}$. Each action specifies the target component or proposed replacement, so that the executor agent can make targeted revisions according to the critic's request.

Valid critic actions are used to construct a constrained executor prompt, from which the executor agent $\mathcal{M}_{\mathrm{exec}}$ proposes up to 4 revised equation skeletons. These candidates first pass a fast screening stage using one split, one restart, a 40-second timeout, and no critic calls. The top candidate, plus the runner-up when its relative score gap is below 0.01, is then evaluated by the full pipeline. We enable early stopping when the training NMSE falls below $10^{-13}$.

In the updating stage, the equation stored in semantic memory comes from the selected candidate after evaluation or, when reflection is triggered, from the comparison between the base candidate and the best refined candidate. Only the better candidate is committed to memory. Before insertion, STRIDE canonicalizes the equation and represents it with term frequency--inverse document frequency (TF-IDF) features over symbolic tokens, including operators, variables, and functions~\citep{salton1988term, ramos2003tfidf}. For a token $t$ in equation $f$, the weight is defined as
\begin{equation}
\label{equation_tfidf}
\mathrm{tfidf}(t,f) = \mathrm{tf}(t,f) \cdot \log\frac{N}{\mathrm{df}(t)},
\end{equation}
where $\mathrm{tf}(t,f)$ denotes the term frequency of $t$ in $f$, $\mathrm{df}(t)$ is the number of equations containing $t$, and $N$ is the total number of equations in the buffer. The similarity between two equations $f_i$ and $f_j$ is then computed using cosine similarity~\citep{manning2008introduction}:
\begin{equation}
\label{equation_sim}
\mathrm{sim}(f_i, f_j) = 
\frac{\mathbf{v}_i \cdot \mathbf{v}_j}
{\|\mathbf{v}_i\| \, \|\mathbf{v}_j\|},
\end{equation}
where $\mathbf{v}_i$ and $\mathbf{v}_j$ are their TF-IDF representations. Equations with cosine similarity above 0.9 are assigned to the same semantic cluster; otherwise, a new cluster is created. This is similar in spirit to diversity-preserving mechanisms in evolutionary symbolic regression~\citep{schmidt2009distilling, udrescu2020aifeynman}. This update rule preserves a representative elite case for each structural region while reducing redundant memory entries for later generator prompts.

\section{Extended Experimental Results}
\subsection{Effect of Data Hints}
\label{sec:effect_data}

We first investigate the effect of data-driven hints in guiding equation discovery. The data hint module extracts lightweight statistical and structural signals from the dataset, including distribution statistics, symmetry properties, and dominant functional patterns.

\begin{figure*}[!t]
    \centering
    \includegraphics[width=0.85\textwidth]{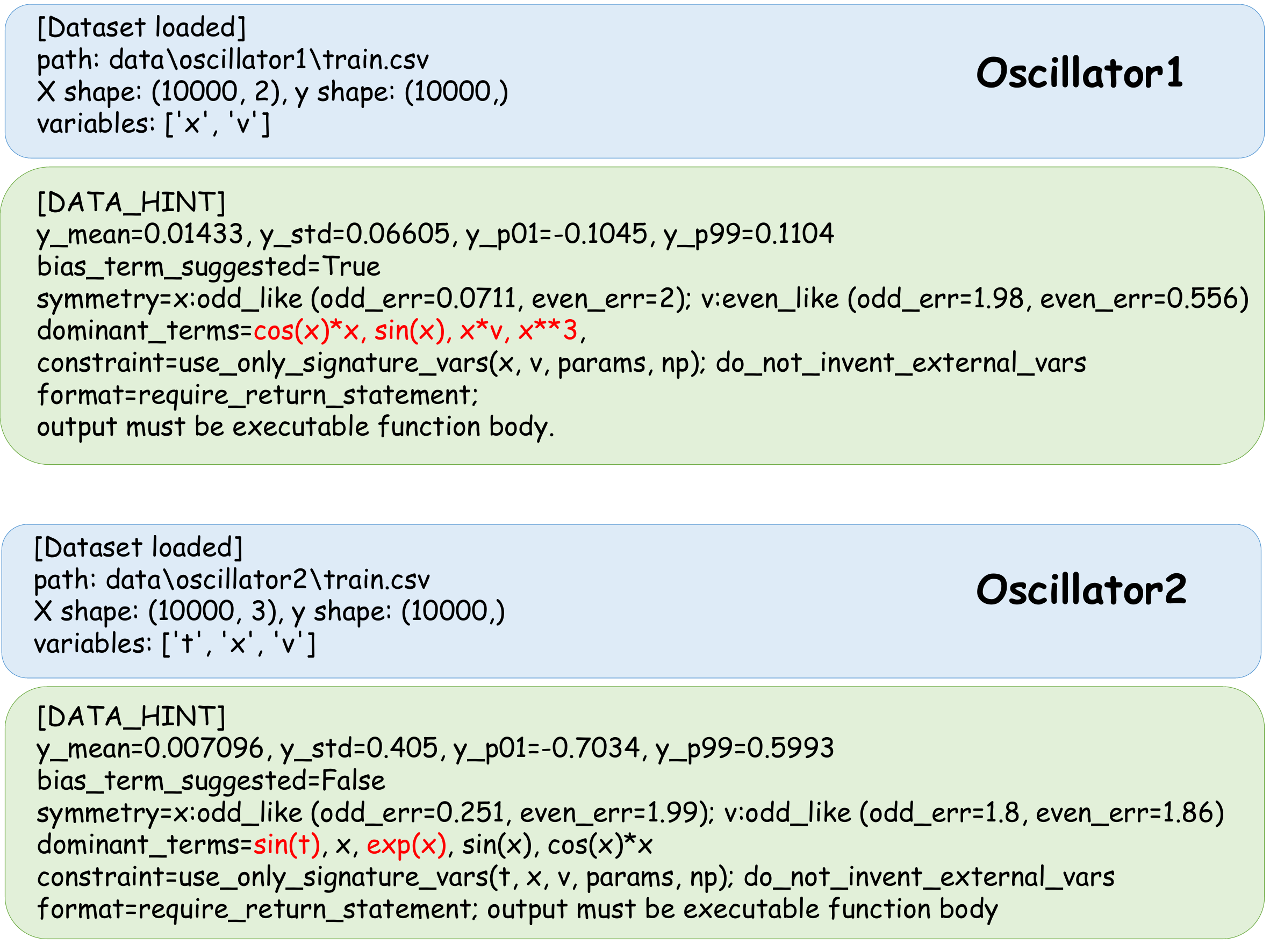}
    \caption{Examples of data hints extracted from training data. The hints summarize distribution statistics, symmetry patterns, dominant terms, and variable constraints.}
    \label{fig:datahint}
\end{figure*}

As shown in Figure~\ref{fig:datahint}, the extracted hints capture key characteristics of the underlying system. For example, in the oscillator datasets, the hints correctly identify symmetry properties (e.g., odd/even behavior in $x$ and $v$) and highlight dominant functional components such as $\sin(x)$, $\cos(x)x$, and interaction terms like $xv$. These signals provide useful inductive bias for the LLM, effectively narrowing the search space while preserving flexibility. 

Importantly, the hints are incorporated as soft guidance rather than hard constraints. This allows the model to balance prior knowledge and exploration, avoiding premature convergence to overly restricted structures. Empirically, we observe that data hints significantly improve early-stage sampling quality and stabilize the search process, leading to faster convergence and better final solutions. Overall, these results demonstrate that data-aware guidance is an effective complement to LLM-based equation discovery.

\subsection{Effect of Mixed Parameter Fitting}
\label{sec:effect_mix}
We further analyze the role of parameter fitting by comparing different fitting strategies, including Gauss--Newton (GN), Levenberg--Marquardt (LM), trust-region methods (TRF), BFGS, and our mixed fitting scheme \citep{more1978lm,nocedal2006numerical,branch1999trf,fletcher1987practical}.

Conventional nonlinear fitting methods treat candidate parameters largely as a coupled optimization vector. GN and LM are often stable for least-squares problems but may fail to achieve high precision when the generated skeleton contains strong nonlinear coupling, while TRF and BFGS can still suffer from sensitivity to initialization or ill-conditioned search landscapes. These limitations make parameter fitting unreliable for complex LLM-generated expressions, and a poor fit can cause a structurally useful skeleton to be underestimated.

\begin{figure}[!tb]
    \centering
    \includegraphics[width=0.4\textwidth]{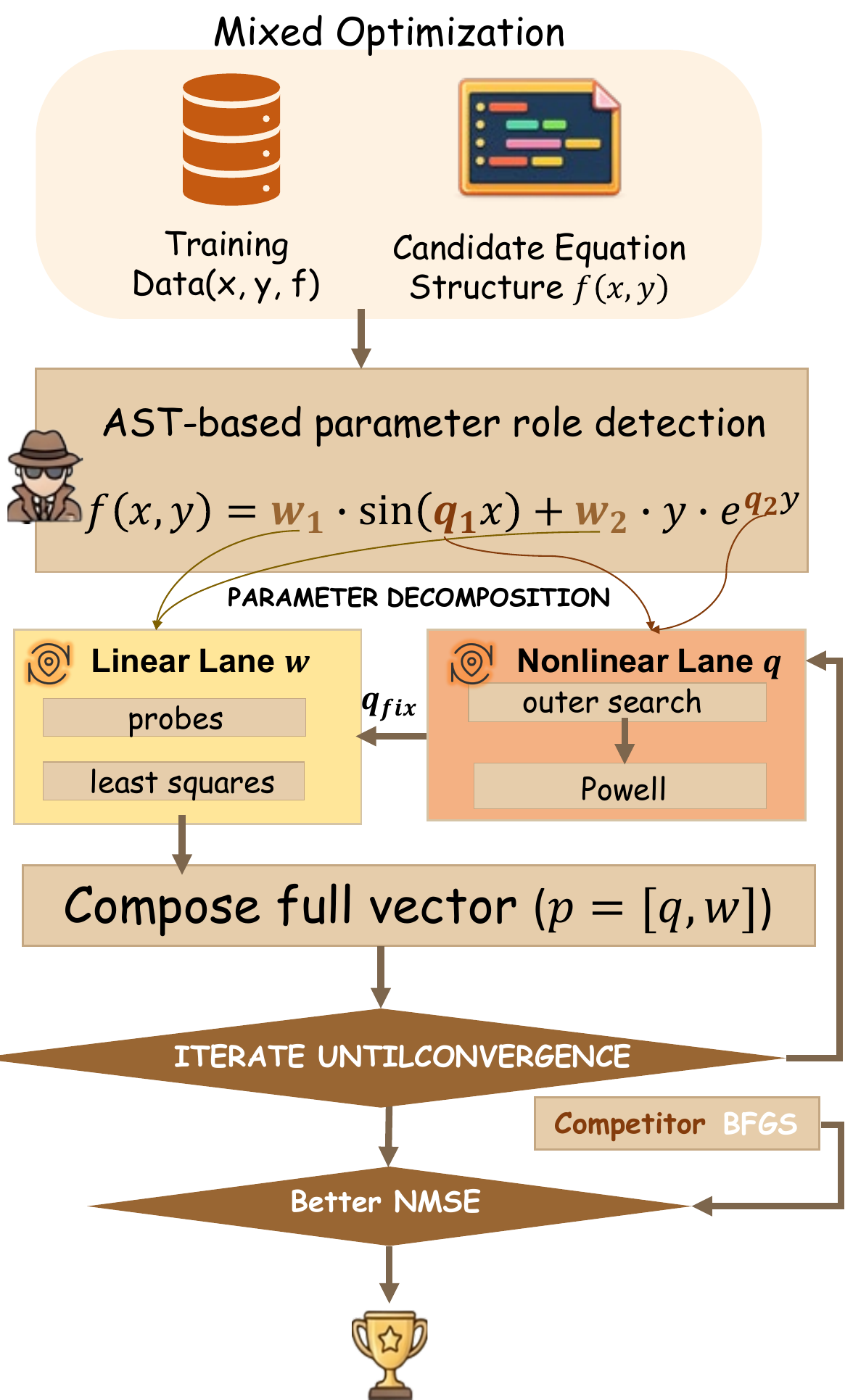}
    \caption{Illustration of the proposed mixed parameter fitting pipeline. Parameters are first decomposed into linear and nonlinear subsets via AST-based role detection. Linear coefficients are solved with least squares, while nonlinear parameters are fitted by multi-start Powell search over the reduced objective. When the decomposition is unreliable for overly complex expressions, BFGS fitting is used as a fallback.}
    \label{fig:mix}
\end{figure}

As illustrated in Figure~\ref{fig:mix}, our mixed fitting method avoids treating all internal parameters as a black box. It first detects parameter roles through AST-based analysis and decomposes them into linear and nonlinear subsets. Under this decomposition, linear coefficients are fitted efficiently by least squares, while nonlinear parameters are fitted through an outer-loop search. This enables the evaluator to use parameter-type information more directly, improving equation precision when the decomposition is effective.

\begin{figure*}[!htbp]
    \centering
    \includegraphics[width=0.92\textwidth]{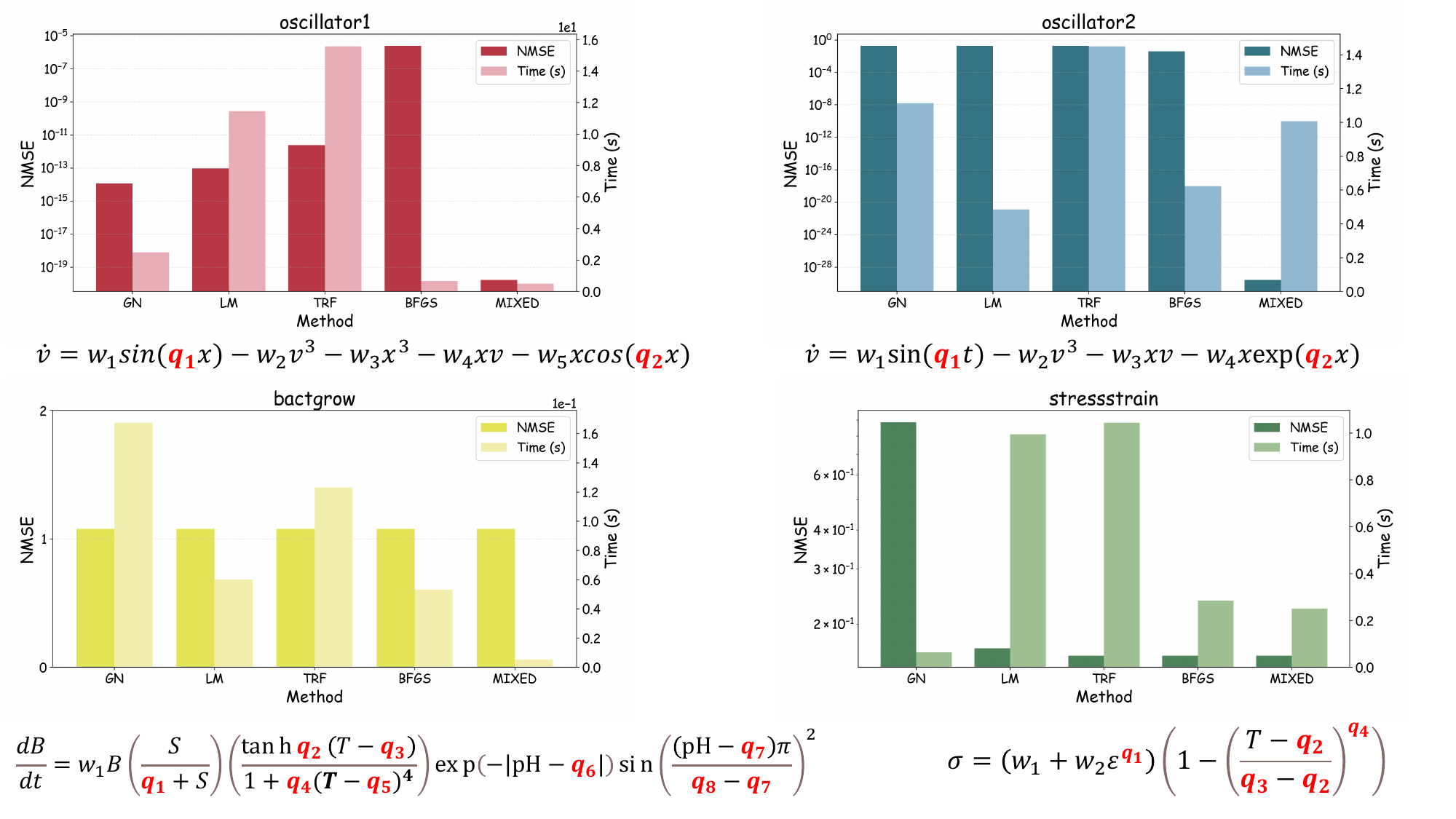}
    \caption{Quantitative comparison of parameter fitting strategies across multiple benchmarks. In the equation legend, \textcolor{red}{red} parameters denote nonlinear parameters.}
    \label{fig:mix_compare}
\end{figure*}

The quantitative comparison in Figure~\ref{fig:mix_compare} shows that the mixed fitting strategy achieves a superior balance between accuracy and computational cost. The results show that this structure-aware fitting strategy gives more reliable parameter estimates across diverse equation forms. We also keep a BFGS fallback for cases where the decomposition is unreliable, which prevents overly complex expressions from failing solely because of imperfect role detection. Importantly, improved parameter fitting directly reduces the risk of discarding structurally correct equation skeletons due to suboptimal coefficient estimation. This highlights that reliable parameter fitting is essential for preserving promising candidates during the search process. This mixed fitting design is closely related to recent efforts that explicitly separate coefficient estimation from structure discovery \citep{negishi2024twostage,zeng2025twostage}.

Overall, mixed fitting strengthens the evaluator as a search signal: candidates are judged by a better-fitted version of their skeletons, so later memory updates and reflection steps are guided by more faithful evidence.

\subsection{Effect of Critic--Executor Reflection}
\label{sec:effect_critic}

We use a critic--executor reflection stage to improve promising equations through constrained symbolic edits.
Given an evaluated candidate with fitted parameters, the critic agent analyzes its structure and fit quality, then proposes local edits from the action space $\{\textsc{Remove},\textsc{Simplify},\textsc{Add}\}$.

Figure~\ref{fig:critic} illustrates the critic--executor reflection pipeline used in the appendix case. After the base expression is fitted and scored, the critic agent diagnoses possible redundant terms, missing components, or unnecessarily complex substructures, and the executor agent instantiates the proposed edits as executable candidates. The restricted action space keeps local repair interpretable and prevents reflection from degenerating into unconstrained regeneration.

The re-evaluation process uses the same mixed fitting and scoring procedure in both stages, but with different evaluation budgets. Refined candidates are first ranked by a lightweight screening pass on a small subset of samples, and only the top candidates are retained. These finalists are then re-evaluated with the full mixed fitting pipeline and compared with the base equation. The higher-scoring equation after this comparison is inserted into the semantic memory buffer.

\subsection{Effect of Semantic Memory}
\label{sec:semantic_memory_effect}
Figure~\ref{fig:cluster_compare} compares TF-IDF-based semantic memory with a score-only retention strategy on \textit{Oscillator 1}. We define high-score candidates as those with Acc$_{0.1}>0.9$. Semantic memory keeps most high-score candidates in singleton or small clusters and suppresses large redundant clusters, indicating that structurally diverse elite cases can be preserved without repeatedly storing the same hypothesis family.

In contrast, score-only retention produces a larger fraction of multi-equation clusters and concentrates high-score candidates within structurally similar groups. These multi-equation clusters also tend to favor shorter expressions, which can repeatedly keep near-duplicate hypotheses while overlooking useful terms carried by more complex but promising equations. This creates two risks for later prompting: redundant experience and premature loss of informative structural components. TF-IDF-based semantic memory mitigates both risks by grouping similar hypotheses, retaining a representative elite case from each cluster, and keeping the retrieved cases structurally varied for later iterations.

\begin{table*}[!t]
    \centering
    \caption[OOD results on symbolic regression and \textsc{LSR-Synth} benchmarks]{OOD results on four symbolic regression benchmarks (top) and four \textsc{LSR-Synth} benchmark suites (bottom). In the top block, Acc$_{0.1}$\textcolor{red!70!black}{$\uparrow$}, Acc$_{0.001}$\textcolor{red!70!black}{$\uparrow$}, and NMSE\textcolor{green!50!black}{$\downarrow$}; in the bottom block, Acc$_{0.1}$\textcolor{red!70!black}{$\uparrow$}, Acc$_{\max,0.1}$\textcolor{red!70!black}{$\uparrow$}, and NMSE\textcolor{green!50!black}{$\downarrow$}. Rows under \textit{Baselines without LLMs} (top) follow the same layout as Table~\ref{tab:main_results} and are shown without ranking highlights; within each LLM backbone block, the best and second-best results are highlighted in \textcolor{red!50}{red} and \textcolor{blue!50}{blue}, respectively.}
    \label{tab:main_results_ood}\label{tab:lsrsynth_results_ood}
    \scriptsize
    \setlength{\tabcolsep}{1pt}
    \renewcommand{\arraystretch}{1.0}
    \resizebox{\textwidth}{!}{
    \begin{tabular}{lcccccccccccc}
    \toprule
    \multirow{2}{*}{\textbf{Method}}
    & \multicolumn{3}{c}{\textbf{Oscillator 1}}
    & \multicolumn{3}{c}{\textbf{Oscillator 2}}
    & \multicolumn{3}{c}{\textbf{E. coli Growth}}
    & \multicolumn{3}{c}{\textbf{Stress-Strain}} \\
    \cmidrule(lr){2-4} \cmidrule(lr){5-7} \cmidrule(lr){8-10} \cmidrule(lr){11-13}
    & {\fontsize{6.2}{6.6}\selectfont Acc$_{0.1}$\textcolor{red!70!black}{$\uparrow$}} & {\fontsize{6.2}{6.6}\selectfont Acc$_{0.001}$\textcolor{red!70!black}{$\uparrow$}} & {\fontsize{6.2}{6.6}\selectfont NMSE\textcolor{green!50!black}{$\downarrow$}}
    & {\fontsize{6.2}{6.6}\selectfont Acc$_{0.1}$\textcolor{red!70!black}{$\uparrow$}} & {\fontsize{6.2}{6.6}\selectfont Acc$_{0.001}$\textcolor{red!70!black}{$\uparrow$}} & {\fontsize{6.2}{6.6}\selectfont NMSE\textcolor{green!50!black}{$\downarrow$}}
    & {\fontsize{6.2}{6.6}\selectfont Acc$_{0.1}$\textcolor{red!70!black}{$\uparrow$}} & {\fontsize{6.2}{6.6}\selectfont Acc$_{0.001}$\textcolor{red!70!black}{$\uparrow$}} & {\fontsize{6.2}{6.6}\selectfont NMSE\textcolor{green!50!black}{$\downarrow$}}
    & {\fontsize{6.2}{6.6}\selectfont Acc$_{0.1}$\textcolor{red!70!black}{$\uparrow$}} & {\fontsize{6.2}{6.6}\selectfont Acc$_{0.001}$\textcolor{red!70!black}{$\uparrow$}} & {\fontsize{6.2}{6.6}\selectfont NMSE\textcolor{green!50!black}{$\downarrow$}} \\
    \midrule
    \rowcolor{gray!10}\multicolumn{13}{l}{\textit{Baselines without LLMs}} \\
    GPlearn & 4.38 & 0.04 & 1.49e{0} & 16.13 & 0.11 & 1.15e{-}1 & 0.76 & 0.01 & 1.00e{0} & 18.97 & 0.27 & 1.11e{0} \\
    PySR & 29.42 & 0.31 & 1.78e{-}1 & 99.39 & 82.10 & 7.88e{-}5 & 3.56 & 0.04 & 3.49e{1} & 87.53 & 0.27 & 6.40e{-}2 \\
    DSR & 3.53 & 0.04 & 7.89e{-}1 & 5.04 & 0.01 & 5.73e{-}1 & 2.97 & 0.04 & 1.02e{0} & 2.98 & 0.00 & 1.15e{0} \\
    uDSR & 3.53 & 0.04 & 7.89e{-}1 & 18.32 & 0.13 & 1.06e{-}1 & 1.51 & 0.01 & 1.04e{0} & 1.90 & 0.00 & 1.52e{0} \\
    \midrule
    \rowcolor{gray!10}\multicolumn{13}{l}{\textit{GPT-5.1}} \\
    LLM-SR & \cellcolor{blue!8}99.97 & \cellcolor{blue!8}41.21 & \cellcolor{blue!8}2.85e{-}4 & \cellcolor{blue!8}86.28 & \cellcolor{blue!8}19.33 & \cellcolor{blue!8}3.05e{-}3 & 0.56 & \cellcolor{blue!8}0.01 & \cellcolor{blue!8}1.20e{0} & 89.02 & 0.14 & \cellcolor{blue!8}8.16e{-}2 \\
    LaSR & 22.48 & 0.21 & 1.12e{-}1 & 2.30 & 0.01 & 8.24e{-}1 & \cellcolor{blue!8}1.18 & 0.01 & 1.42e{1} & 88.89 & 0.14 & 9.86e{-}2 \\
    SR-LLM & 46.10 & 0.21 & 1.08e{-}2 & 16.21 & 0.12 & 1.15e{-}1 & 0.66 & 0.01 & 1.28e{0} & 26.02 & 0.14 & 4.68e{-}1 \\
    DrSR & 65.33 & 3.78 & 3.35e{-}2 & 47.36 & 0.16 & 4.49e{-}2 & 0.51 & 0.01 & 2.40e{1} & \cellcolor{blue!8}90.65 & \cellcolor{red!8}2.98 & \cellcolor{blue!8}7.02e{-}2 \\
    STRIDE & \cellcolor{red!8}100.00 & \cellcolor{red!8}100.00 & \cellcolor{red!8}5.97e{-}12 & \cellcolor{red!8}100.00 & \cellcolor{red!8}100.00 & \cellcolor{red!8}4.48e{-}21 & \cellcolor{red!8}4.58 & \cellcolor{red!8}0.05 & \cellcolor{red!8}1.41e{-}1 & \cellcolor{red!8}93.77 & \cellcolor{blue!8}0.41 & \cellcolor{red!8}4.79e{-}2 \\
    \midrule
    \rowcolor{gray!10}\multicolumn{13}{l}{\textit{Claude-3-Haiku}} \\
    LLM-SR & \cellcolor{blue!8}67.43 & \cellcolor{blue!8}8.77 & \cellcolor{blue!8}3.01e{-}2 & \cellcolor{red!8}100.00 & \cellcolor{blue!8}93.84 & \cellcolor{blue!8}1.40e{-}5 & \cellcolor{blue!8}4.75 & \cellcolor{red!8}0.07 & \cellcolor{red!8}5.03e{-}3 & 9.62 & 0.14 & 3.22e{-}1 \\
    LaSR & 28.39 & 0.39 & 1.20e{-}1 & 76.29 & 2.96 & 1.02e{-}2 & 1.07 & 0.01 & 3.43e{+}0 & \cellcolor{blue!8}88.35 & \cellcolor{blue!8}0.41 & \cellcolor{blue!8}7.91e{-}2 \\
    SR-LLM & 4.45 & 0.04 & 1.61e{+}3 & 16.21 & 0.12 & 1.15e{-}1 & 0.99 & 0.02 & 3.00e{+}0 & 73.44 & 0.14 & 2.75e{-}1 \\
    DrSR & 66.81 & 0.70 & 3.37e{-}2 & 40.82 & 0.45 & 4.93e{-}2 & 0.59 & 0.00 & 4.75e{+}0 & 30.08 & 0.00 & 3.07e{-}1 \\
    \textbf{STRIDE} & \cellcolor{red!8}100.00 & \cellcolor{red!8}100.00 & \cellcolor{red!8}7.82e{-}8 & \cellcolor{red!8}100.00 & \cellcolor{red!8}100.00 & \cellcolor{red!8}4.66e{-}8 & \cellcolor{red!8}9.23 & \cellcolor{red!8}0.07 & \cellcolor{red!8}5.03e{-}3 & \cellcolor{red!8}92.82 & \cellcolor{red!8}9.08 & \cellcolor{red!8}5.50e{-}2 \\
    \midrule
    \midrule
    \multirow{2}{*}{\textbf{Method}}
    & \multicolumn{3}{c}{\textbf{CRK}}
    & \multicolumn{3}{c}{\textbf{PO}}
    & \multicolumn{3}{c}{\textbf{MatSci}}
    & \multicolumn{3}{c}{\textbf{BPG}} \\
    \cmidrule(lr){2-4} \cmidrule(lr){5-7} \cmidrule(lr){8-10} \cmidrule(lr){11-13}
    & {\fontsize{6.0}{6.4}\selectfont Acc$_{0.1}$\textcolor{red!70!black}{$\uparrow$}} & {\fontsize{6.0}{6.4}\selectfont Acc$_{\max,0.1}$\textcolor{red!70!black}{$\uparrow$}} & {\fontsize{6.0}{6.4}\selectfont NMSE\textcolor{green!50!black}{$\downarrow$}}
    & {\fontsize{6.0}{6.4}\selectfont Acc$_{0.1}$\textcolor{red!70!black}{$\uparrow$}} & {\fontsize{6.0}{6.4}\selectfont Acc$_{\max,0.1}$\textcolor{red!70!black}{$\uparrow$}} & {\fontsize{6.0}{6.4}\selectfont NMSE\textcolor{green!50!black}{$\downarrow$}}
    & {\fontsize{6.0}{6.4}\selectfont Acc$_{0.1}$\textcolor{red!70!black}{$\uparrow$}} & {\fontsize{6.0}{6.4}\selectfont Acc$_{\max,0.1}$\textcolor{red!70!black}{$\uparrow$}} & {\fontsize{6.0}{6.4}\selectfont NMSE\textcolor{green!50!black}{$\downarrow$}}
    & {\fontsize{6.0}{6.4}\selectfont Acc$_{0.1}$\textcolor{red!70!black}{$\uparrow$}} & {\fontsize{6.0}{6.4}\selectfont Acc$_{\max,0.1}$\textcolor{red!70!black}{$\uparrow$}} & {\fontsize{6.0}{6.4}\selectfont NMSE\textcolor{green!50!black}{$\downarrow$}} \\
    \midrule
    \rowcolor{gray!10}\multicolumn{13}{l}{\textit{GPT-5.1}} \\
    LLM-SR & 65.40 & 40.00 & 4.20e{-}1 & 64.96 & 0.00 & 7.43e{2} & \cellcolor{red!8}100.00 & \cellcolor{red!8}100.00 & \cellcolor{red!8}1.11e{-}1 & 60.64 & 60.00 & 6.94e{0} \\
    SR-Scientist & 42.32 & \cellcolor{blue!8}40.00 & 3.59e{0} & 44.48 & 0.00 & \cellcolor{red!8}1.31e{-}2 & \cellcolor{red!8}100.00 & \cellcolor{red!8}100.00 & \cellcolor{blue!8}2.11e{-}1 & 47.56 & 20.00 & \cellcolor{blue!8}8.99e{-}1 \\
    \textbf{STRIDE} & \cellcolor{red!8}100.00 & \cellcolor{red!8}100.00 & \cellcolor{red!8}1.46e{-}3 & \cellcolor{blue!8}57.56 & 0.00 & \cellcolor{blue!8}4.77e{-}2 & \cellcolor{red!8}100.00 & \cellcolor{red!8}100.00 & 2.50e{-}1 & \cellcolor{red!8}80.00 & \cellcolor{red!8}80.00 & \cellcolor{red!8}5.09e{-}2 \\
    \midrule
    \rowcolor{gray!10}\multicolumn{13}{l}{\textit{Claude-3-Haiku}} \\
    LLM-SR & 41.56 & \cellcolor{blue!8}40.00 & 1.55e{1} & 48.60 & 0.00 & 2.02e{5} & 80.00 & \cellcolor{blue!8}80.00 & 5.12e{0} & \cellcolor{blue!8}44.84 & 0.00 & 2.12e{2} \\
    SR-Scientist & \cellcolor{blue!8}72.36 & 20.00 & \cellcolor{blue!8}8.39e{-}1 & \cellcolor{red!8}75.00 & 0.00 & \cellcolor{blue!8}2.99e{-}2 & \cellcolor{red!8}100.00 & \cellcolor{red!8}100.00 & \cellcolor{blue!8}1.84e{-}1 & 42.28 & \cellcolor{blue!8}20.00 & \cellcolor{blue!8}1.38e{2} \\
    \textbf{STRIDE} & \cellcolor{red!8}100.00 & \cellcolor{red!8}100.00 & \cellcolor{red!8}5.52e{-}3 & \cellcolor{blue!8}60.12 & 0.00 & \cellcolor{red!8}1.06e{-}2 & \cellcolor{red!8}100.00 & \cellcolor{red!8}100.00 & \cellcolor{red!8}1.76e{-}1 & \cellcolor{red!8}96.60 & \cellcolor{red!8}60.00 & \cellcolor{red!8}2.46e{-}4 \\
    \bottomrule
    \end{tabular}
    }
    \end{table*}

\subsection{Extended OOD Analysis}
\label{app:ood_results}

Table~\ref{tab:main_results_ood} reports additional out-of-distribution (OOD) results across both the main benchmarks and \textsc{LSR-Synth} suites. Compared to in-distribution performance, OOD results exhibit larger variance across methods, highlighting the importance of structural generalization rather than pure fitting accuracy.

From the upper panel of Table~\ref{tab:main_results_ood}, classical methods and several LLM-based baselines suffer significant degradation in NMSE under distribution shift, particularly on \textit{E. coli Growth} and \textit{Stress-Strain}, where complex nonlinear interactions dominate. In contrast, our method maintains consistently low NMSE and high Acc$_{0.1}$ across all tasks, indicating improved robustness to unseen regimes.
The lower panel further shows that this advantage extends to the more diverse \textsc{LSR-Synth} benchmarks: while competing methods are often unstable across suites, our approach achieves the best or near-best results in most cases, especially on CRK and BPG, where precise parameter estimation is critical.

Overall, these results indicate that STRIDE's multi-role self-reflective workflow does more than improve local fitting quality: reliable mixed fitting, targeted critic--executor repair, and semantic memory coordinate across iterations to preserve symbolic structures that remain meaningful under distribution shift.

\section{Baseline Implementation Details}
\label{app:baseline_details}

We compare our method with both traditional symbolic regression baselines and recent LLM-based equation discovery methods.
To ensure a fair comparison, we keep the evaluation data splits fixed across all methods.
For LLM-based baselines, we evaluate both \textsc{GPT-5.1} and \textsc{Claude-3-Haiku} using the same prompt template and decoding temperature as in the main experiments, unless a comparison is explicitly reported for a single backbone.

\paragraph{gplearn~\citep{stephens2022gplearn}.}
\texttt{gplearn} is a classical genetic-programming SR baseline that evolves symbolic expression trees through selection, crossover, and mutation over a predefined operator set. We use the public \texttt{gplearn} package with population size 1000, 2000 generations, and tournament size 20.

\paragraph{PySR~\citep{cranmer2023pysr}.}
\texttt{PySR} is an advanced SR method that employs asynchronous multi-island GP-based evolution with built-in constant optimization. We use \texttt{PySR} as a strong evolutionary baseline and run 2000 search iterations with 50 populations.

\paragraph{DSR and uDSR~\citep{petersen2021dsr,landajuela2022udsr}.}
\texttt{DSR} is a neural-guided symbolic regression method that learns a policy for sampling expressions, while \texttt{uDSR} extends this line by integrating multiple symbolic regression strategies into a unified modular framework. For \texttt{DSR}, we use the original deep symbolic regression framework based on neural sequence generation and policy-gradient optimization, with training batch size 100 and 200000 sampled expressions. For \texttt{uDSR}, we keep the same batch size and sample budget, and additionally enable its polynomial, constant, length-prior, and polynomial-optimizer settings.

\paragraph{LLM-SR~\citep{shojaee2025llmsr}.}
\texttt{LLM-SR} is an LLM-guided program-search method that represents equations as executable programs and improves them through iterative proposal, evaluation, and island-based memory updating. We implement \texttt{LLM-SR} following its original formulation with \texttt{global\_max\_sample\_num}=2500, \texttt{samples\_per\_prompt}=4, and \texttt{num\_islands}=10.

\paragraph{LaSR~\citep{grayeli2024lasr}.}
\texttt{LaSR} is a concept-library-based LLM-SR method that uses abstract natural-language concepts distilled from high-performing hypotheses to guide subsequent symbolic search. We implement \texttt{LaSR} with 20 populations and 100 iterations, using LLM operation probability 0.01, 5 generated equations, 3 generated concepts, and 2 concept-crossover samples.

\paragraph{SR-LLM~\citep{guo2025srllm}.}
\texttt{SR-LLM} is a retrieval-augmented incremental symbolic regression method that reuses prior symbolic experience to guide later equation generation. We retain its core retrieval-and-generation workflow and, following the released setting summary, use batch size 500, 20 epochs, and 30 evolution rounds, corresponding to about 300000 explored samples.

\paragraph{SR-Scientist~\citep{xia2026srscientist}.}
\texttt{SR-Scientist} is an agentic equation-discovery baseline that combines tool-assisted data analysis with iterative equation proposal and evaluation. We implement \texttt{SR-Scientist} following its original agentic workflow. To keep the comparison computationally comparable, we use the same two-backbone protocol; its main inference parameters are \texttt{num\_turns}=100, \texttt{max\_assistant\_turns}=25, \texttt{mape\_threshold}=0.001, and \texttt{top\_k}=3.

\paragraph{DrSR~\citep{wang2025drsr}.}
\texttt{DrSR} is a dual-reasoning LLM-based SR method that combines data-aware analysis with reflective symbolic search. We preserve its core data-understanding and reflection mechanisms while using the same benchmark setup and two-backbone protocol as the other LLM-based baselines. The main search parameters are 700 iterations, 4 samples per prompt, and 10 islands.

\section{Dataset Descriptions}
\label{app:dataset}

To evaluate the effectiveness and generalization ability of our method, we conduct experiments on a collection of benchmark tasks drawn from two primary sources: the datasets introduced in the original \textsc{LLM-SR} work and the \textsc{LSR-Synth} benchmark suite. These datasets cover a diverse range of scientific domains, including physics, biology, chemistry, and materials science. Most tasks are associated with known underlying equations that serve as ground-truth targets for symbolic regression. The following paragraphs summarize the key characteristics of these benchmarks, and Table~\ref{tab:llmsr_benchmarks} then lists the corresponding equations in the same local context.
\subsection{LLM-SR Benchmarks}
\label{app:datasets_llmsr}

The \textsc{LLM-SR} benchmarks consist of four representative symbolic regression tasks designed to evaluate the ability of models to recover nontrivial scientific relationships from data; their equations are listed in Table~\ref{tab:llmsr_benchmarks}(a). These tasks include two nonlinear oscillator systems, a biological growth process, and a temperature-dependent stress-strain model. The oscillator and growth tasks are defined by explicit governing equations. For \textit{Stress-Strain}, which does not have a single standardized ground-truth law, we report the representative form used to generate and describe the benchmark behavior. The oscillator systems emphasize dynamic interactions between position and velocity, while the growth and material models incorporate nonlinear responses and environmental dependencies. Together, these benchmarks provide a controlled yet diverse testbed for assessing equation discovery performance.

\subsection{LSR-Synth Benchmark Suites}

We categorize equation difficulty in the \textsc{LSR-Synth} benchmark using the ranking script in our benchmark pipeline. The score combines expression tree depth, operator diversity, variable coupling, inverse or non-smooth operators (e.g., $\log$, $\sqrt{\cdot}$, division), domain-related terms, and empirical penalties. We rank equations within each suite by this score and select the top five tasks for evaluation, as listed in Table~\ref{tab:llmsr_benchmarks}(b). The domains are denoted as \textbf{CRK}, \textbf{BPG}, \textbf{PO}, and \textbf{MatSci}, following the categorization in \textit{LLM-SRBench}. Each selected task uses the benchmark-provided training, in-domain test, and out-of-domain test splits.

We observe that higher-ranked equations tend to exhibit stronger nonlinear coupling and non-smooth operators, which significantly increases optimization difficulty and leads to lower discovery success rates. Table~\ref{tab:llmsr_benchmarks} summarizes these benchmark equations immediately below: the upper panel lists the four \textsc{LLM-SR} tasks, and the lower panel lists the selected hard \textsc{LSR-Synth} equations.

\begin{table*}[!t]
\centering
\caption{Dataset equations used in our experiments. The upper panel summarizes the four \textsc{LLM-SR} benchmark tasks, and the lower panel lists the top five hardest \textsc{LSR-Synth} equations in each domain according to the proposed difficulty score. The equation identifiers follow the local zero-based benchmark files, while the displayed formulas are matched to the corresponding one-based entries in the \textsc{LLM-SRBench} appendix.}
\label{tab:llmsr_benchmarks}\label{tab:hardest_equations}
\small
\textbf{(a) Four \textsc{LLM-SR} benchmark tasks.}

\setlength{\tabcolsep}{3pt}
\renewcommand{\arraystretch}{1.15}
\resizebox{\textwidth}{!}{
\begin{tabular}{@{}p{2.0cm} p{1.6cm} p{11.0cm}@{}}
\toprule
\textbf{Dataset} & \textbf{Inputs} & \textbf{Underlying or representative equation} \\
\midrule

\textbf{Oscillator 1}
& $x,v$
& $\dot{v}=F\sin(\omega x)-\alpha v^3-\beta x^3-\gamma xv-x\cos(x)$
\\[2pt]

\textbf{Oscillator 2}
& $t,x,v$
& $\dot{v}=F\sin(\omega t)-\alpha v^3-\beta xv-\delta x\exp(\gamma x)$
\\[2pt]

\textbf{Bactgrow}
& $B,S,T,\mathrm{pH}$
& \resizebox{\linewidth}{!}{$\displaystyle
\frac{dB}{dt}=
\mu_{\max}B\!\left(\frac{S}{K_S+S}\right)
\left(\frac{\tanh k(T-x_0)}{1+c(T-x_{\mathrm{decay}})^4}\right)
\exp\!\left(-|\mathrm{pH}-\mathrm{pH}_{\mathrm{opt}}|\right)
\sin\!\left(\left(\frac{\mathrm{pH}-\mathrm{pH}_{\min}}
{\mathrm{pH}_{\max}-\mathrm{pH}_{\min}}\pi\right)^2\right)
$}
\\[4pt]

\textbf{Stress-Strain}
& $\varepsilon,T$
& $\sigma=(A+B\varepsilon^n)\left(1-\left(\frac{T-T_r}{T_m-T_r}\right)^m\right)$ \quad (representative form)
\\

\bottomrule
\end{tabular}
}

\textbf{(b) Top five hardest \textsc{LSR-Synth} equations in each domain.}

\small
\setlength{\tabcolsep}{6pt}
\renewcommand{\arraystretch}{1.15}
\resizebox{\textwidth}{!}{
\begin{tabular}{l l l}
\toprule
\textbf{Domain} & \textbf{Equation ID} & \textbf{Equation} \\
\midrule

\multirow{5}{*}{Chemistry}
& CRK22 & $\dfrac{dA}{dt} = -kA(t)^2 - kA(t)\exp(-k_s t) + \dfrac{k_z A(t)^2}{\beta A(t)^4 + 1}$ \\

& CRK29 & $\dfrac{dA}{dt} = -kA(t) - kA(t)\exp(-k_s t) + \dfrac{k_z A(t)^2}{\beta A(t)^4 + 1}$ \\

& CRK3  & $\dfrac{dA}{dt} = -kA(t)^2 - kA(t)\exp(-k_s t) + k_w \cos(\log(A(t)+1))$ \\

& CRK15 & $\dfrac{dA}{dt} = -k\sqrt{A(t)} - kA(t)\exp(-k_s t) + k_p \sin(\log(A(t)+1))$ \\

& CRK19 & $\dfrac{dA}{dt} = -kA(t)^2 - kA(t)\exp(-k_s t) + k_p \sin(\log(A(t)+1))$ \\

\midrule

\multirow{5}{*}{Biology}
& BPG20 & $\dfrac{dP}{dt} =
r\left(-1+\dfrac{P(t)}{\alpha}\right)\left(1-\dfrac{P(t)}{K_0}\right)P(t)
+ r\left(1-\dfrac{P(t)}{K_0}\right)P(t)
+ \dfrac{rP(t)}{1+\exp\!\left(-\alpha(-\beta+P(t))\right)}$ \\

& BPG21 & $\dfrac{dP}{dt} =
r\left(-1+\dfrac{P(t)}{\alpha}\right)\left(1-\dfrac{P(t)}{K_0}\right)P(t)
+ \dfrac{rP(t)}{t^{\alpha}}$ \\

& BPG13 & $\dfrac{dP}{dt} =
r\left(-1+\dfrac{P(t)}{\alpha}\right)\left(1-\dfrac{P(t)}{K_0}\right)P(t)
+ rP(t)
+ \dfrac{rP(t)}{1+\exp\!\left(-\alpha(-\beta+P(t))\right)}$ \\

& BPG9  & $\dfrac{dP}{dt} =
r\left(-1+\dfrac{P(t)}{\alpha}\right)\left(1-\dfrac{P(t)}{K_0}\right)P(t)
+ r\left(1-\dfrac{P(t)}{K_0}\right)P(t)
+ r\left(1-\exp(-\gamma P(t))\right)P(t)$ \\

& BPG19 & $\dfrac{dP}{dt} = r\left(1-\dfrac{P(t)}{K_0}\right)P(t) + \dfrac{rP(t)}{t^{\alpha}}$ \\

\midrule

\multirow{5}{*}{Physics}
& PO20 & $\dfrac{dv}{dt} = -\alpha v(t)^3 - \beta\log(|v(t)|+1) - 2\beta v(t) - \mu(1-v(t)^2)v(t) - \omega_0^2(\gamma |v(t)|^{0.33}+1)x(t)$ \\

& PO40 & $\dfrac{dv}{dt} = F_0\sin(t) - \beta |v(t)|^{0.33} - \omega_0^2(\gamma |v(t)|^{0.33}+1)x(t) - \omega_0^2 x(t)^3 - \omega_0^2 x(t)$ \\

& PO12 & $\dfrac{dv}{dt} = F_0\sin(t) - \alpha v(t)^3 - \beta |v(t)|^{0.33} - \omega_0^2(\gamma t+1)x(t) - \omega_0^2 x(t)$ \\

& PO5  & $\dfrac{dv}{dt} = -\beta\sin(v(t)) - 2\beta v(t) - \omega_0^2(\gamma |v(t)|^{0.33}+1)x(t) - \omega_0^2 x(t)^3 - \omega_0^2 x(t)$ \\

& PO14 & $\dfrac{dv}{dt} = F_0\sin(t) - \beta\log(|v(t)|+1) - \beta\sin(v(t)) - 2\beta v(t) - \mu(1-x(t)^2)v(t)$ \\

\midrule

\multirow{5}{*}{Material}
& MatSci12 & $\sigma = E_0\epsilon(-\alpha_T(T-T_0)+1) + K\epsilon^N\exp\!\left(-\dfrac{Q}{RT}\right) + \eta\epsilon\exp(-(T-T_0)^2)$ \\

& MatSci23 & $\sigma = K\epsilon^N\exp\!\left(-\dfrac{Q}{RT}\right) - \beta(T-T_0) + \eta(T-T_0)\log(\epsilon+1)$ \\

& MatSci5  & $\sigma = E_0\epsilon(-\alpha_T(T-T_0)+1) + K\epsilon^N\exp\!\left(-\dfrac{Q}{RT}\right) + \eta\epsilon^M(T-T_0)$ \\

& MatSci1  & $\sigma = H\epsilon^3 + K\epsilon^N\exp\!\left(-\dfrac{Q}{RT}\right) + \eta\epsilon\sin(T-T_0)$ \\

& MatSci3  & $\sigma = H\epsilon^3 + K\epsilon^N\exp\!\left(-\dfrac{Q}{RT}\right) + \eta\epsilon^3(T-T_0)$ \\

\bottomrule
\end{tabular}
}
\end{table*}

\section{Additional Case Study}
\label{app:case_study}
\begin{figure*}[!p]
    \centering
    \begin{subfigure}{0.98\textwidth}
        \centering
        \includegraphics[width=\linewidth]{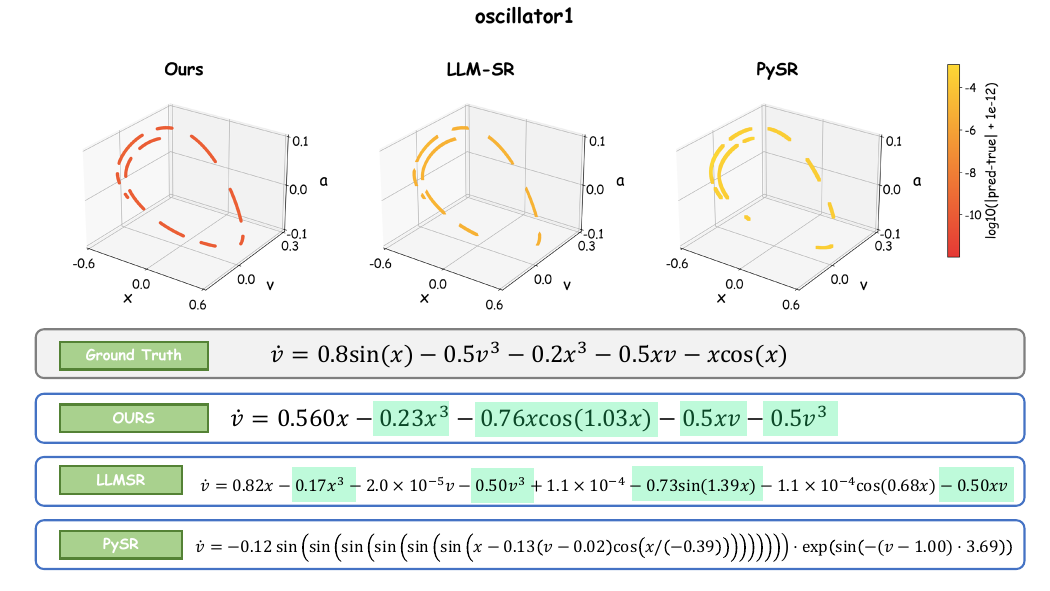}
        \caption{\textit{Oscillator 1}.}
        \label{fig:osc1_case}
    \end{subfigure}

    \begin{subfigure}{0.98\textwidth}
        \centering
        \includegraphics[width=\linewidth]{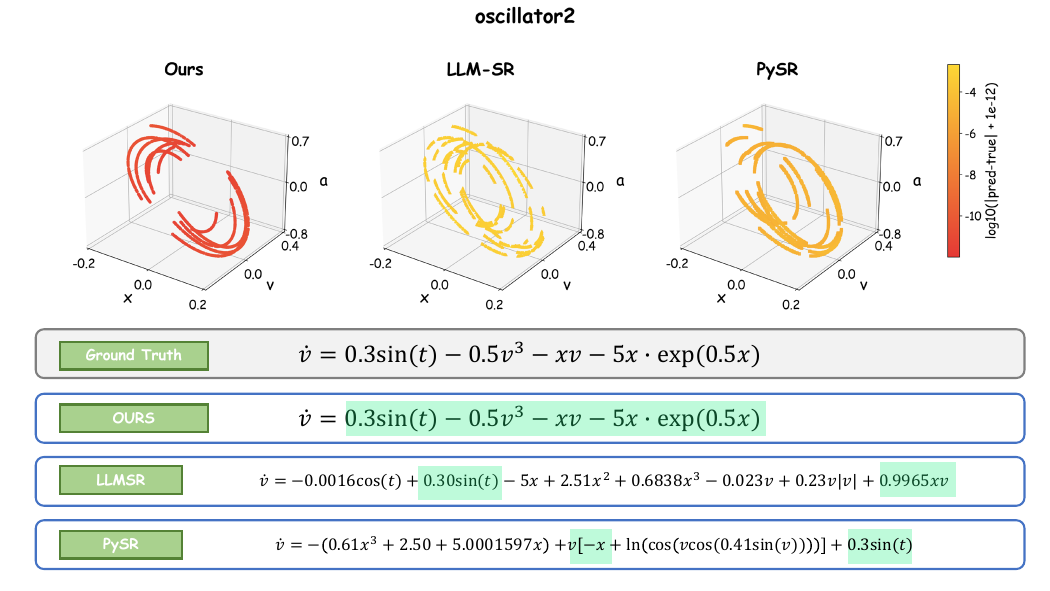}
        \caption{\textit{Oscillator 2}.}
        \label{fig:osc2_case}
    \end{subfigure}
    \caption{Case studies on the two oscillator benchmarks. We compare phase-space overlays and recovered expressions for \textbf{Ours}, \textbf{LLM-SR}, and \textbf{PySR}. Highlighted terms indicate components aligned with the ground-truth equation.}
    \label{fig:oscillator_cases}
\end{figure*}

We use \textit{Oscillator 1} as a qualitative case because it requires recovering a compact nonlinear dynamic law rather than only fitting a smooth response surface. The task contains coupled terms involving position and velocity, so methods that add many weak or redundant terms may still reduce pointwise error locally but fail to preserve the correct phase-space structure.
As shown in Figure~\ref{fig:osc1_case}, STRIDE produces the closest phase-space alignment with the ground-truth acceleration field. This behavior is consistent with the equation-level evidence: the discovered expression retains the main nonlinear interaction patterns of the oscillator, including coupled position-velocity effects and nonlinear restoring terms. In contrast, the competing expressions either miss part of the symbolic structure or compensate with more complex terms that do not align as well with the true dynamics. Thus, the accuracy gain in this case is better explained by improved symbolic identification than by merely accumulating additional polynomial terms.

Figure~\ref{fig:osc2_case} provides a second oscillator case. Here, STRIDE recovers the dominant forcing, cubic damping, velocity-position coupling, and exponential restoring term, leading to a phase-space trajectory that closely follows the ground truth. LLM-SR captures several relevant components but introduces extra polynomial and absolute-value terms, while PySR recovers partial forcing information but misses important structural interactions. This case again suggests that the improvement comes from identifying the correct symbolic components and their couplings.

\begin{figure*}[!p]
    \centering
    \includegraphics[width=0.86\textwidth]{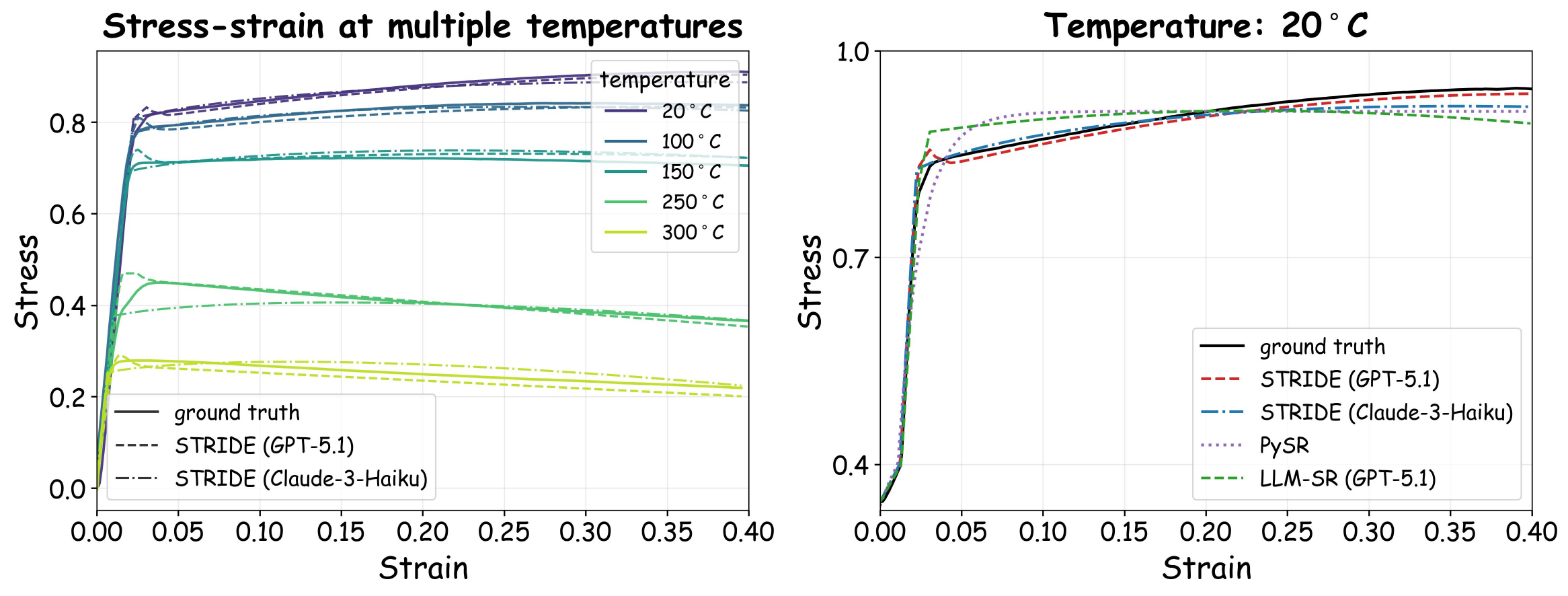}
    \caption{Case study on \textit{Stress-Strain}. The left panel compares predictions across multiple temperatures, and the right panel zooms into the 20$^\circ$C condition.}
    \label{fig:stress_case}
\end{figure*}

Figure~\ref{fig:stress_case} shows the stress-strain case under multiple temperatures. STRIDE tracks the ground-truth curves across temperature conditions more consistently than the compared methods, especially in the rapid transition region at small strain and in the high-strain saturation regime. This indicates that the recovered equation captures the temperature-dependent response pattern rather than only fitting one local curve.

\begin{figure*}[!p]
    \centering
    \includegraphics[width=\textwidth]{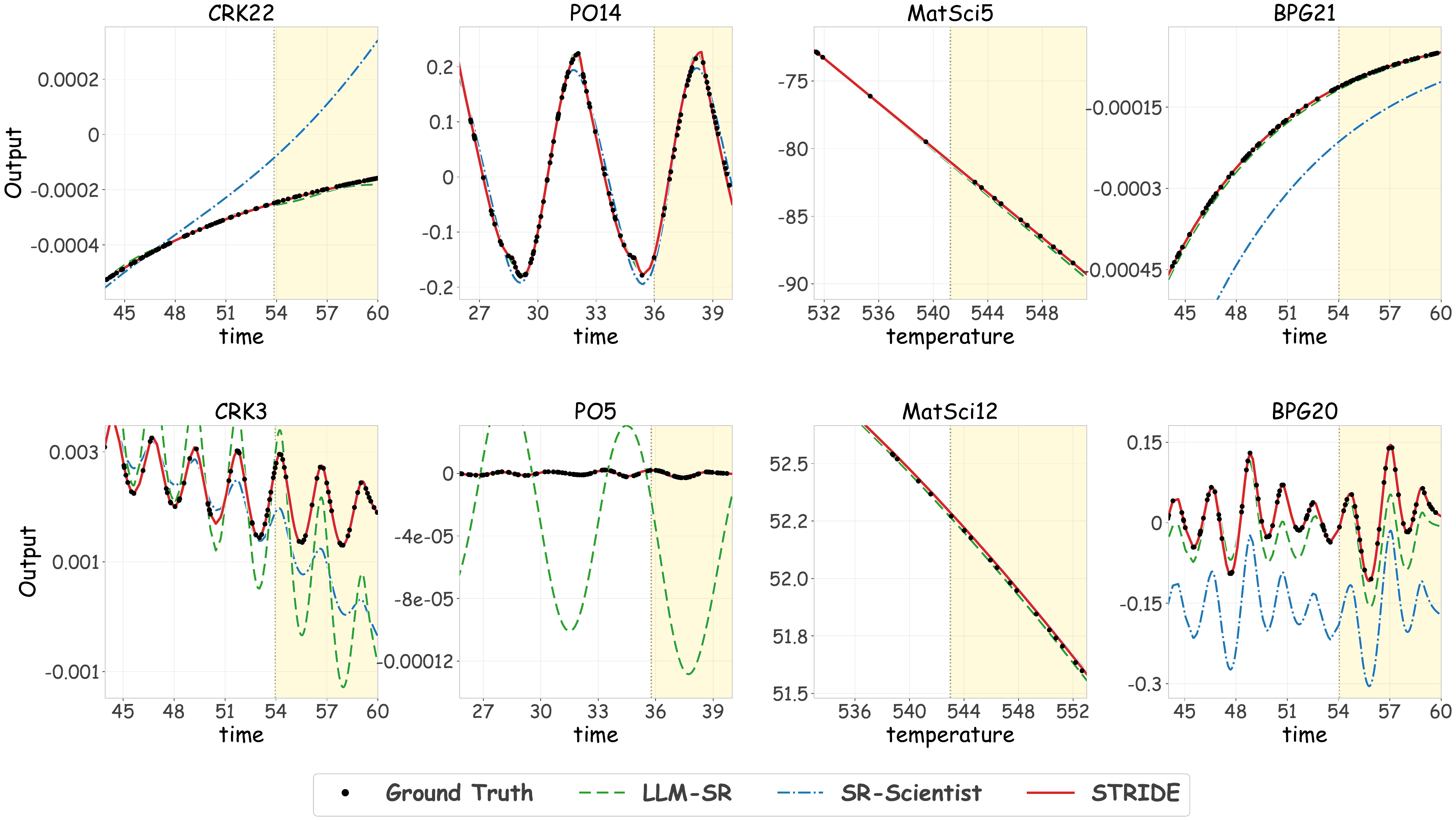}
    \caption{Representative LSR-Synth case studies. The shaded regions denote OOD intervals, and the curves compare ground truth, LLM-SR, SR-Scientist, and STRIDE. The upper row uses GPT-5.1 as the backbone, and the lower row uses Claude-3-Haiku.}
    \label{fig:lsr_case}
\end{figure*}

Finally, Figure~\ref{fig:lsr_case} visualizes representative LSR-Synth cases from CRK, PO, MatSci, and BPG. The shaded regions indicate extrapolation intervals. Across these examples, STRIDE generally stays closer to the ground-truth trajectories in both observed and extrapolated regions, whereas competing methods often deviate after the transition into the OOD interval. These cases qualitatively support the OOD results in Appendix~\ref{app:ood_results}: the selected equations are more likely to preserve the underlying functional structure needed for extrapolation.

\section{Prompt Templates}

\begin{figure*}[p]
    \centering

    \begin{subfigure}{0.88\textwidth}
        \centering
        \includegraphics[width=\linewidth]{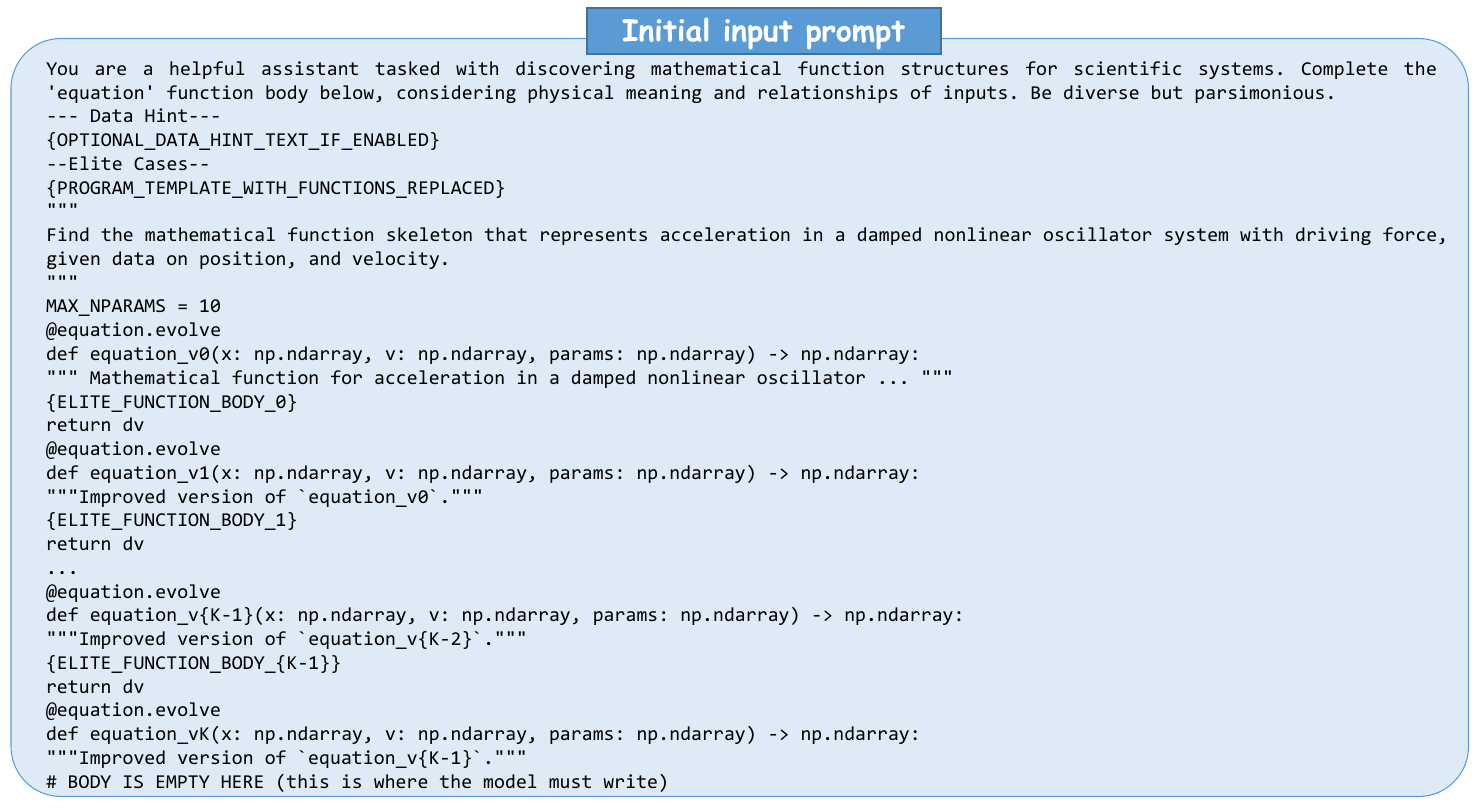}
        \caption{Prompt for the generator agent.}
        \label{fig:initialprompt}
    \end{subfigure}

    \begin{subfigure}{0.88\textwidth}
        \centering
        \includegraphics[width=\linewidth]{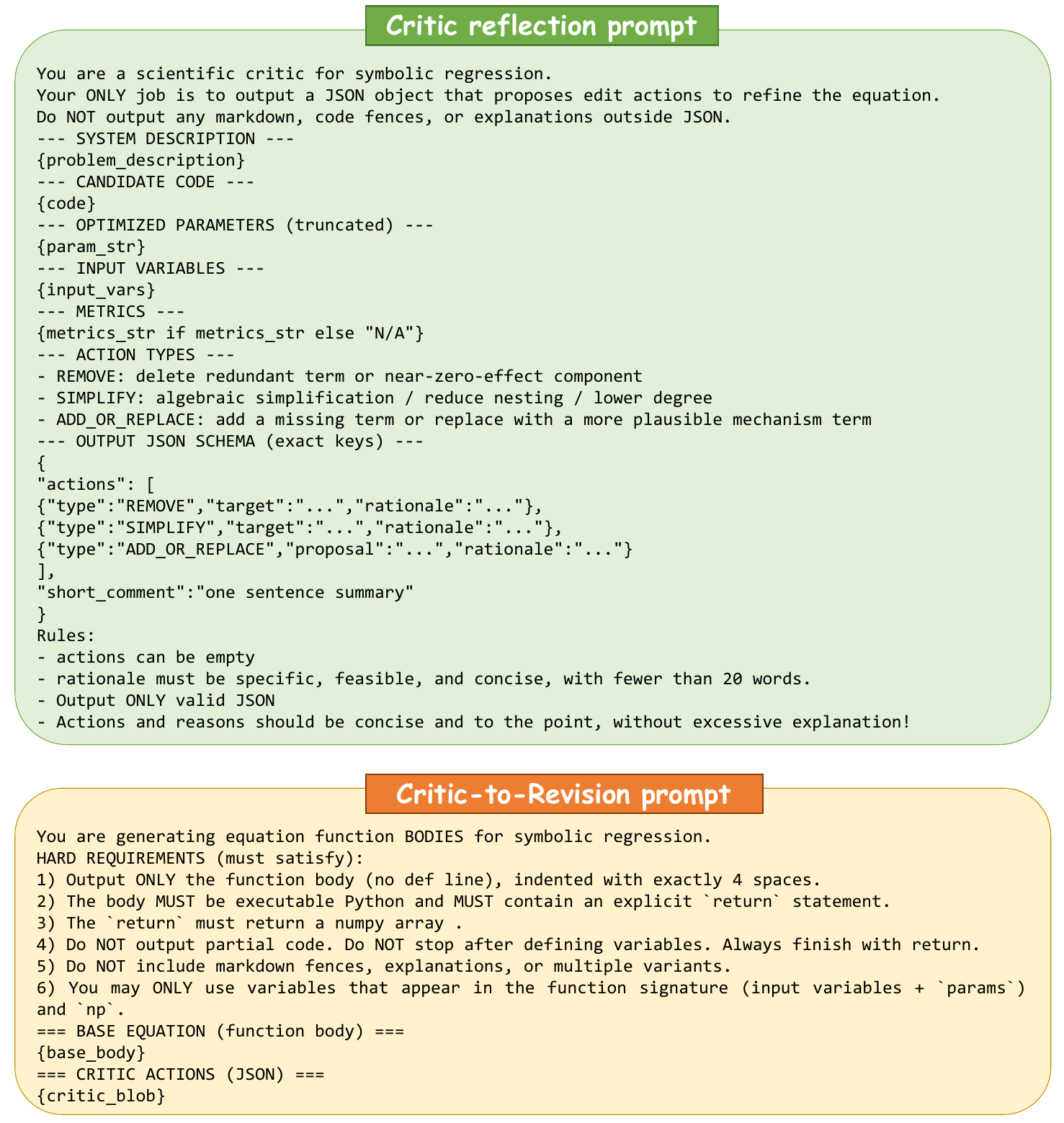}
        \caption{Prompt for the critic and executor agents.}
        \label{fig:criticprompt}
    \end{subfigure}

    \caption{Prompt templates for the generator agent and critic--executor agents.}
    \label{fig:prompt_all}
\end{figure*}

The prompt design follows the multi-role workflow in Section~\ref{sec:method}. Figure~\ref{fig:initialprompt} gives the generator agent task context, variables, data hints, and retrieved elite cases for proposing executable skeletons. Figure~\ref{fig:criticprompt} gives the critic--executor agents the evaluated candidate, fitted parameters, score feedback, context, and constrained actions for diagnosing weaknesses and producing targeted revisions.
\begin{figure*}[!p]
    \centering
    \includegraphics[width=0.82\textwidth]{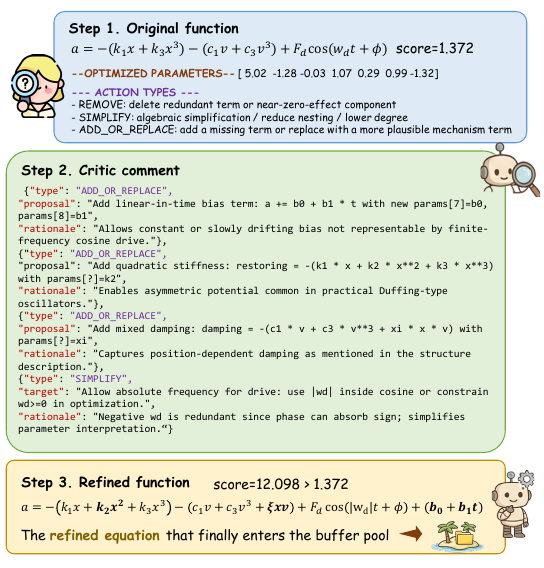}
    \caption{Illustration of critic--executor reflective repair. The critic agent analyzes a fitted candidate equation, proposes constrained symbolic edits, and the executor agent turns valid actions into locally repaired candidates.}
    \label{fig:critic}
\end{figure*}

\end{document}